\documentclass[preprint,12pt]{elsarticle}

\usepackage{amsmath,amssymb}
\usepackage{graphicx}
\usepackage{booktabs}
\usepackage{tabularx}
\usepackage{xcolor}
\usepackage{float}
\usepackage{algorithm}
\usepackage{algpseudocode}
\usepackage{rotating}
\usepackage{pdflscape}
\usepackage{url}
\biboptions{sort&compress}

\begin{document}

\begin{frontmatter}

\title{BAT-NO: A Boundary-Condition-Aware Transformer Neural Operator for Crashworthiness Prediction of Vehicle Components}

\author[label1]{Haoran Li}
\author[label1]{Yingxue Zhao}
\author[label1]{Haosu Zhou}
\author[label2]{Mustapha Ziane}
\author[label2]{Pierre Culiere}
\author[label3]{Tobias Pfaff}
\author[label1]{Nan Li\corref{cor}}

\cortext[cor]{Corresponding author}
\ead{n.li09@imperial.ac.uk}

\affiliation[label1]{organization={Dyson School of Design Engineering, Imperial College London},
            city={London},
            country={UK}}

\affiliation[label2]{organization={Keysight Technologies},
            city={Bagneux},
            country={France}}

\affiliation[label3]{organization={Prometheus},
            country={UK}}

\begin{abstract}
High-fidelity finite-element simulations provide accurate crashworthiness predictions, but their cost limits iterative design exploration. While deep learning surrogate models can reduce this cost, many existing component-level models are developed under a single prescribed boundary condition, which may limit their generalisability to variations in boundary conditions. This work proposes a Boundary-Condition-Aware Transformer Neural Operator (BAT-NO) for autoregressive prediction of transient displacement fields and scalar crashworthiness responses under simultaneous variations in geometry and boundary conditions. A B-pillar simulation framework is developed to evaluate the model's generalisation across variations in geometry, impact position and velocity, and support stiffness. BAT-NO combines recurrent mesh processing and latent-grid Fourier operator processing. Boundary condition information is transferred to the latent grid through a hybrid local--global mechanism. Slice-based attention models interactions among physically related regions, while direct boundary-to-grid projection preserves local spatial structure. Across the validation sets for the shape-only, shape-and-loading, and shape-loading-boundary cases, BAT-NO achieves the lowest mean final-step mean nodal Euclidean displacement error among the evaluated baselines. In the most challenging validation case, it reduces the mean error by 32.6\% relative to the second-best model. Hyperparameter tuning further reduces the validation error from $0.451$ to $0.269$~mm, with a comparable error of $0.267$~mm on 300 unseen test simulations sampled within the investigated design space. An attention-based scalar decoder jointly predicts six response trajectories with a mean relative error of 2.46\%. Most derived crashworthiness indicators have median errors below 3\%. These results show that explicit local and global boundary-condition representations improve crashworthiness prediction over expanded component-level design spaces.
\end{abstract}

\begin{keyword}
Crashworthiness analysis \sep Neural operators \sep Graph neural networks \sep Surrogate modelling \sep Boundary conditions
\end{keyword}

\end{frontmatter}

\section{Introduction}
\label{sec:introduction}

Crashworthiness analysis is a crucial task in vehicle structural design. It requires the prediction of transient deformation, contact response, energy absorption, reaction forces, and intrusion-related quantities under impact loading. High-fidelity finite-element (FE) simulations remain the standard tool for this task because of their high accuracy in predicting structural responses involving large deformation, nonlinear material behaviour, and complex contact interactions. However, repeated FE simulations are computationally expensive, especially for design exploration or optimisation tasks which require iterative simulation.

Neural-network surrogate models have therefore been increasingly studied for crashworthiness prediction. These models aim to approximate the response of high-fidelity simulations at a lower computational cost. Image-based convolutional neural network (CNN) surrogates have been applied to physical-field prediction and manufacturability assessment in sheet forming \cite{zhou2022imagestamping,attar2021rapid,attar2021hfqcnn,xu2022supermeshing}. Related learned representations have supported constitutive-response prediction and geometry optimisation \cite{ding2024cnnconstitutive,attar2023implicit,attar2023stamping}. CNN has also been used to predict the crashworthiness performance of vehicle panel components \cite{li2024cnncrash}. Graph neural networks (GNNs) provide a natural representation for mesh-based physical simulation and have been increasingly applied across mechanics-related domains \cite{zhao2024gnnreview}. In graph-based learned simulators for mesh, nodes represent mesh nodes and edges represent nodal connectivities. MeshGraphNet (MGN) \cite{pfaff2021meshgraphnets} introduced autoregressive physical simulation using message passing over dynamically constructed graphs with an encoder-processor-decoder architecture. MultiScale MeshGraphNets \cite{fortunato2022multiscale} and X-MeshGraphNet \cite{nabian2024xmeshgraphnet} further improve long-range information transfer and scalability through coarse graph levels or partitioned graph processing. Related hierarchical graph surrogates have been developed for sheet-forming simulations, including a multi-level GCNN with image-inspired spatial edge weights \cite{zhou2025multilevelgcnn} and RUGNN, which combines recurrent temporal modelling with graph-based downsampling and upsampling \cite{zhao2026rugnn,zhao2025batterybox}. These models provide strong foundations for mesh-based field prediction, but they remain primarily based on local message passing. For large structural meshes, long-range load transfer may require many processor steps or explicit multiscale construction.

GNN-based simulators operate on a discretised mesh and learn mappings tied to its nodes and connectivity. Neural operators provide a complementary formulation by learning mappings between functions, allowing physical inputs and outputs to be defined as fields over a continuous domain rather than solely through a particular mesh discretisation. This operator perspective can therefore support evaluation on different discretisations while treating geometry, displacement state, loading conditions, and boundary conditions as input functions. DeepONet \cite{lu2021deeponet} represents an operator through branch and trunk networks. The Fourier Neural Operator (FNO) \cite{li2021fno} learns global convolutional operators in the Fourier domain on regular grids. The Graph Neural Operator (GNO) \cite{li2020gno} extends kernel integral operators to irregular point sets and meshes. Geometry-Informed Neural Operator (GINO) \cite{li2023gino} combines GNO-based geometry-grid projection with FNO processing on a latent grid, enabling operator learning on non-Cartesian geometries. Transformer-based operator models, including Transolver \cite{wu2024transolver} and Universal Physics Transformers (UPT) \cite{alkin2024upt}, further improve global communication through learned physical slices or latent tokens. These architectures are relevant for crashworthiness because they provide mechanisms for modelling long-range interactions and for handling varying input functions.

Recent studies have begun to apply graph and operator models to industrial crash dynamics. Li et al. \cite{li2026regunet} proposed Recurrent Graph U-Net (ReGUNet) for B-pillar crash deformation prediction, and showed that hierarchical graph processing with recurrent hidden-state propagation can improve autoregressive displacement-field prediction. Mask-Morph Graph U-Net further addressed crash-field prediction under large geometric variation through geometry-aware coarse-graph morphing and masked pretraining \cite{li2026maskmorph}. Nabian et al. \cite{nabian2026automotive} applied Transolver to automotive crash analysis and compared it with MGN, demonstrating the potential of attention-based neural operators for modelling complex crash dynamics. GeoTransolver \cite{adams2025geotransolver} introduced geometry-aware attention for operator learning on complex irregular domains. Akhare et al. \cite{akhare2026crash} subsequently applied GeoTransolver to bumper-beam and full-vehicle crash dynamics and introduced GeoTS-FLARE, a low-rank attention variant that approximately halves the attention-block memory overhead while improving crash-field prediction. MeshGraphNet-Transformer \cite{iparraguirre2026mgnt} is a hybrid local--global model that combines local graph message passing with attention-based global feature updates. Curtosi et al. \cite{curtosi2026crash} investigated MeshTransolver, MeshGeoTransolver, and MeshGeoFLARE for full-vehicle lateral pole-impact prediction. These architectures combine local mesh message passing with global physics-aware or geometry-aware attention, while a sparse contact-aware variant models dynamic proximity interactions during autoregressive rollout. CarCrashNet \cite{elrefaie2026carcrashnet} provides a large-scale benchmark containing component- and vehicle-level crash simulations, field trajectories, scalar histories, and broader design-parameter variation. These studies establish the feasibility of learned crash surrogates for industrial-scale transient dynamics.

Existing crash-surrogate studies have demonstrated component-level prediction under fixed test conditions and vehicle-level prediction on large industrial datasets. However, fewer studies consider how changes in neighbouring structures may be reflected in the conditions transmitted through a component's connection regions. In this study, this assembly-level influence is approximated at the component level by varying the support stiffnesses at these regions. Standard crashworthiness tests are usually defined under fixed loading and boundary conditions, so geometry-only surrogate modelling is a valid and useful setting for exploring component design. However, in an assembled structure, the response of a component is strongly affected by its connections to neighbouring parts. Changes in adjacent components or joint stiffness are transmitted through the connection regions and appear at the component level as changes in boundary conditions. A component surrogate trained only under a single fixed boundary configuration may therefore have limited ability to represent the response of the same component when embedded in different structural environments. This motivates a component-level formulation in which the surrogate learns not only the effect of geometry, but also the effect of change in boundary conditions. In this setting, boundary and loading information should be represented explicitly and spatially, while the model should still preserve local mesh fidelity and capture global structural interactions.

This work proposes Boundary-Condition-Aware Transformer Neural Operator (BAT-NO) for crashworthiness prediction. The model is designed to learn transient physical fields for components with varying shapes, loading and support conditions. It combines local mesh-state processing with latent-grid FNO processing. The mesh processor captures local structural state and edge information. Following the idea of GINO \cite{li2023gino}, a GNO kernel-based mesh-to-grid projection maps the processed mesh representation to a regular latent grid, where global Fourier operator processing is performed. To represent spatially heterogeneous conditions, BAT-NO introduces a boundary-condition-aware slice pathway and a direct boundary-to-grid projection. Together, these pathways form a hybrid local--global projection. The direct projection preserves spatially local support, contact, and loading information, while the slice pathway, inspired by Transolver \cite{wu2024transolver}, captures their global relationships through slice attention before field evolution is modelled on the latent grid. The processed latent grid features are then reconstructed back to mesh domain with similar kernel-based layer, followed by the output decoder.

In addition to field prediction, BAT-NO includes an attention-based scalar decoder. The decoder extracts relevant information from the learned representation and predicts global and local crashworthiness responses. During fully coupled rollout, the predicted scalar responses can be fed back into the next time step together with the predicted displacement field. This allows the surrogate to reconstruct the deformation trajectory and estimate engineering response quantities within one architecture.

The main contributions of this work are as follows:
\begin{enumerate}
    \item A component-level crash simulation framework is formulated to emulate the transmission of boundary-condition changes from neighbouring structures by systematically varying the support parameters.
    \item A boundary-condition-aware surrogate model, BAT-NO, is proposed with a hybrid local--global boundary-condition processing module, in which slice-based attention captures global interactions among physically related regions and direct boundary-to-grid projection preserves local spatial information.
    \item A reusable attention-based scalar decoder is developed to jointly predict deformation fields and scalar crashworthiness responses during autoregressive rollout, and can be attached to different surrogate model architectures when scalar-response prediction is required.
    \item The proposed BAT-NO is systematically evaluated against representative GNN, neural-operator, and transformer-based baselines, with ablation studies used to quantify the contribution of each architectural component.
\end{enumerate}

\section{Related Work}
\label{sec:related_work}

\subsection{Neural Operators for Physical Field Prediction}
\label{subsec:related_neural_operators}

Neural operators learn mappings between function spaces and have become a widely used surrogate-modelling framework for physical field prediction. Unlike conventional neural networks that learn a mapping between fixed-dimensional vectors, neural operators are designed to approximate solution operators of parameterised physical systems. This makes them attractive for engineering simulations, where the objective is often to predict a spatial field under changes in geometry, material parameters, loading, or boundary conditions.

FNO \cite{li2021fno} is one of the most established neural operator architectures. FNO first lifts the input field to a latent feature space, applies repeated Fourier layers, and then projects the latent representation back to the target field. In each Fourier layer, the feature field is transformed to the spectral domain via the fast Fourier transform (FFT), a truncated set of Fourier modes is multiplied by learned complex-valued weights, and the result is transformed back to the physical domain via the inverse FFT. This provides an efficient global receptive field and allows the learned operator to be evaluated at different grid resolutions. However, the standard FNO formulation assumes that the input and output fields are represented on regular grids. This limits its direct application to FE crash simulations, where the structural domain is represented by irregular meshes with varying geometry. The spectral truncation in FNO may also limit the resolution of sharply localised deformation and other physical responses. Several extensions have been proposed to improve its applicability, including U-FNO \cite{wen2022ufno}, which combines Fourier layers with U-Net-type local paths for multiscale flow prediction. Another example is Geo-FNO \cite{li2022geofno}, which learns a deformation from an irregular physical domain to a regular latent domain where Fourier layers can be applied.

GNO \cite{li2020gno} extends neural operator learning to irregular point sets and meshes by discretising kernel integral operators over graph neighbourhoods. In GNO, each node corresponds to a point where the physical field is evaluated, and information from neighbouring points is aggregated to approximate a learned integral operator on the domain. GNO therefore provides a natural mechanism for operator learning on unstructured meshes without first interpolating the data to a regular grid. This is important for FE domains, where nodal position and connectivity carry direct physical information. The main limitation is that local graph neighbourhoods require multiple layers to communicate information over long distances, while dense kernel connections are expensive for large three-dimensional (3D) meshes. For crashworthiness prediction, this creates a trade-off between local geometric fidelity and efficient modelling of long-range structural load transfer.

GINO \cite{li2023gino} combines the geometric flexibility of GNO \cite{li2020gno} with the global modelling efficiency of FNO \cite{li2021fno}. A GNO encoder maps information from an irregular mesh or point cloud to a regular latent grid, an FNO processes the latent grid, and a GNO decoder queries the predicted field back on the target geometry. This design allows global spectral processing while retaining the ability to handle non-Cartesian domains. GINO also uses geometric information, such as signed-distance fields, to describe the relationship between the latent grid and the physical domain. Recent variants extend this idea in different directions. GINOT \cite{liu2025ginot} replaces parts of the geometry encoding and decoding pipeline with transformer attention, allowing geometry information from unordered point clouds to be fused with solution queries. EqGINO \cite{kim2026eqgino} further introduces equivariant GNO and FNO components to improve consistency under rotations and translations in 3D partial differential equation problems. These developments show that GINO-type architectures are effective templates for combining irregular geometry handling with latent-grid operator learning. However, in crashworthiness and many other structural dynamic applications, the loading and boundary conditions are spatially heterogeneous and physically distinct. Treating these quantities only as generic input channels may be insufficient when contact regions, supports, and evolving scalar responses strongly affect the deformation trajectory.

Transformer-based neural operators provide another route to global communication on irregular domains. Transolver \cite{wu2024transolver} introduces Physics-Attention, which replaces attention over all mesh points with attention over a smaller set of learned physical-state slices. Each point is softly assigned to latent slices according to its features. The points assigned to each slice are aggregated into slice tokens, self-attention is applied between the tokens, and the updated information is mapped back to the points. For a smaller number of slices, this reduces the dependence on mesh size compared with full pointwise attention while still allowing long-range information exchange. This is relevant for crashworthiness because different structural regions may participate in common deformation or load-transfer modes even when they are spatially separated. Transolver++ \cite{luo2025transolverpp} improves the scalability of this formulation for million-scale geometries by introducing more efficient parallelism and adaptive mechanisms. More recent members of the Transolver family further target very large industrial geometries \cite{wu2026transolver3}. These models demonstrate the value of tokenising physical states rather than directly attending over all mesh nodes. Nevertheless, the original Transolver formulation does not explicitly decompose evolving structural state, geometry, boundary constraints, and loading effects. For crash simulations with varying support and impact conditions, this motivates architectures that use slice-based global communication while preserving explicit boundary conditioning.

\subsection{Boundary-Condition-Aware Neural Operators}
\label{subsec:related_bc_neural_operators}

Several recent studies have investigated neural operators that explicitly represent variable boundary conditions. These methods differ from standard operator-learning settings where all samples share the same boundary configuration and the boundary effect can be absorbed implicitly into the training distribution.

A closely related direction is to learn boundary-to-domain function extensions. Mousavi et al. \cite{mousavi2026boundary} proposed a general framework for imposing non-homogeneous boundary conditions on neural operators by mapping boundary data to latent pseudo-fields over the full spatial domain. The extended boundary representation can then be consumed by a standard neural operator. This boundary-extension perspective motivates the idea of boundary-to-grid projection developed in this work, which is introduced in Section~\ref{sec:method}.

GeoTransolver \cite{adams2025geotransolver} introduces geometry, global properties, and boundary-condition context into a transformer-based neural operator for irregular domains. Its geometry-aware attention mechanism couples learned state slices with a shared context representation computed from multiscale geometric queries. This provides a relevant example of using attention to condition physics surrogates on geometry and operating conditions. Akhare et al. \cite{akhare2026crash} further introduce GeoTS-FLARE, a memory-efficient GeoTransolver variant that retains geometry-aware cross-attention while replacing slice-based self-attention with low-rank latent routing. They apply the model to high-fidelity bumper-beam and full-vehicle crash datasets, reporting an approximately twofold reduction in attention-block memory overhead while accurately predicting plastic deformation patterns and acceleration profiles at critical occupant locations.

Lifting Product Fourier Neural Operator (LP-FNO) \cite{kashi2024lpfno} studies the boundary-to-domain setting directly by learning mappings from lower-dimensional boundary functions to full-domain solution fields. It applies FNO to boundary data and lifts the resulting representation into the solution domain through a lifting-product layer. This provides a structured operator-learning approach for spatially varying boundary inputs. Physics-Informed Transformer Neural Operator (PINTO) \cite{boya2025pinto} provides another relevant example of boundary-aware conditioning through attention. It uses cross-attention to modulate space-time coordinate representations with initial/boundary points and values, allowing the model to generalise to unseen initial and boundary conditions. Finally, arbitrary-domain neural operators such as Region Interacting Graph Neural Operator (RIGNO) \cite{mousavi2025rigno} and Geometry Aware Operator Transformer (GAOT) \cite{wen2025gaot} provide useful context for learning on irregular geometries. RIGNO uses a regional graph-based representation and multiscale message passing to learn solution operators on arbitrary point clouds. GAOT combines multiscale attentional graph neural operator encoders and decoders with geometry embeddings and a transformer processor. Both methods address geometric generalisation more directly than standard grid-based neural operators. Although they can incorporate boundary information, their primary architectural emphasis is arbitrary-domain operator learning rather than explicit conditioning on varying support, contact, and loading fields.

Overall, existing boundary-aware neural operators show that boundary conditions should be represented as structured inputs rather than only as global auxiliary parameters. The remaining challenge is to incorporate such information into a transient crashworthiness surrogate, where irregular mesh geometry, local contact, support conditions, loading magnitude, and autoregressive response history jointly determine the deformation field. BAT-NO addresses this setting by combining local mesh-state processing, boundary-condition-aware slice projection, direct boundary-to-grid projection, and global latent-grid operator learning.

\section{Problem Formulation}
\label{sec:problem_formulation}

Crashworthiness analysis of vehicle panel components involves nonlinear transient solid dynamics with large deformation, contact, material nonlinearity, and time-dependent loading. At a continuum level, such problems can be viewed as dynamic initial-boundary-value problems governed by the balance of linear momentum,
\begin{equation}
    \rho\,\ddot{\mathbf{u}}
    =
    \nabla_{\mathbf{x}} \cdot \boldsymbol{\sigma}
    +
    \rho\,\mathbf{f}_{\mathrm b},
    \qquad
    \mathbf{x}\in\Omega(t),
\end{equation}
together with initial conditions, boundary conditions, loading conditions, contact interactions, and constitutive relations. Here, $\rho$, $\mathbf{u}$, $\boldsymbol{\sigma}$, and $\mathbf{f}_{\mathrm b}$ are spatial- and time-dependent fields denoting the density, displacement, Cauchy stress tensor, and body force per unit mass, respectively. Based on this physical setting, we simplify and formulate surrogate modelling as learning a temporal response operator for crash dynamics. Such an operator is not tied to a single spatial discretisation and can, in principle, be evaluated on meshes, point clouds, or query locations in the physical domain. Although this formulation can in principle support different spatial representations, the present study evaluates BAT-NO only on the FE mesh discretisation used to generate the datasets. We use a graph representation to retain local geometric information, by encoding quantities that may be associated with graph edges. The objective is to map the current structural state, geometry, and boundary condition to the next-step response.

Each B-pillar component at time step $t$ is represented by a mesh graph
\begin{equation}
    \mathcal{G}_{t} =
    \left(
    \mathcal{V},
    \mathcal{E}
    \right),
\end{equation}
where $\mathcal{V}$ is the set of mesh nodes and $\mathcal{E}$ is the set of mesh edges. For node $i\in\mathcal{V}$, the nodal coordinate is denoted by $\mathbf{x}_{i,t}\in\mathbb{R}^{3}$. The displacement state is denoted by $\mathbf{u}_{i,t}\in\mathbb{R}^{3}$ and the incremental displacement at time step $t$ is defined as
\begin{equation}
    \label{eq:incremental-displacement}
    \Delta \mathbf{u}_{i,t}
    =
    \mathbf{u}_{i,t}
    -
    \mathbf{u}_{i,t-1}.
\end{equation}
The node and edge features over the graph are denoted by $\mathbf{V}_{t}=\{\mathbf{v}_{i,t}\}_{i\in\mathcal{V}}$ and $\mathbf{E}_{t}=\{\mathbf{e}_{ij,t}\}_{(i,j)\in\mathcal{E}}$, respectively. The nodal structural-state feature is denoted by $\mathbf{v}_{i,t}$ and contains the current incremental displacement and nodal coordinate,
\begin{equation}
    \mathbf{v}_{i,t}
    =
    \left[
    \Delta\mathbf{u}_{i,t},\,
    \mathbf{x}_{i,t}
    \right].
\end{equation}
The boundary and loading information associated with node $i$ is represented separately by the condition feature $\mathbf{b}_{i,t}$. The corresponding stacked feature matrices are denoted by $\mathbf{B}_{t}=\{\mathbf{b}_{i,t}\}_{i\in\mathcal{V}}$. The detailed construction of the boundary-condition features is described in Section~\ref{sec:datasets_experiments}.

For an edge $(i,j)\in\mathcal{E}$, the edge feature contains the relative geometry in both the reference and current configurations,
\begin{equation}
    \mathbf{e}_{ij,t}
    =
    \left[
    \mathbf{r}^{0}_{ij},\,
    \left\|\mathbf{r}^{0}_{ij}\right\|_{2},\,
    \mathbf{r}_{ij,t},\,
    \left\|\mathbf{r}_{ij,t}\right\|_{2}
    \right]
    \in\mathbb{R}^{8},
\end{equation}
where
\begin{equation}
    \mathbf{r}^{0}_{ij}
    =
    \mathbf{x}_{i,0}
    -
    \mathbf{x}_{j,0},
    \qquad
    \mathbf{r}_{ij,t}
    =
    \mathbf{x}_{i,t}
    -
    \mathbf{x}_{j,t}.
\end{equation}
Here, $\left\|\mathbf{r}_{ij,t}\right\|_{2}$ denotes the Euclidean distance between nodes $i$ and $j$ at time $t$. The reference-geometry channels remain fixed throughout the trajectory, while the current-geometry channels are recomputed from the predicted nodal coordinates during autoregressive rollout. For each undirected mesh edge, both directed edges $(i,j)$ and $(j,i)$ are included.

Using the node and edge representations defined above, the primary objective is to predict the transient deformation field of the component. In this work, the field target is the nodal incremental displacement, denoted by $\Delta\mathbf{U}_{t}$. By accumulating the predicted incremental displacements over time, the deformed component shape at the final crash state can be obtained. In addition to field prediction, the model can also predict global scalar-response quantities for more detailed crashworthiness analysis. Let $\mathbf{s}_{t}\in\mathbb{R}^{d_s}$ denote the scalar-response vector at time step $t$, including quantities such as energy, reaction force, contact force, or intrusion.

The one-step prediction task is written as
\begin{equation}
    \left(
    \Delta\widehat{\mathbf{U}}_{t+1},
    \widehat{\mathbf{s}}_{t+1},
    \mathbf{H}^{r}_{t+1}
    \right)
    =
    \mathcal{O}_{\theta}
    \left(
    \mathbf{X}_{t},
    \Delta\mathbf{U}_{t},
    \mathbf{B}_{t},
    \mathbf{E}_{t},
    \mathbf{s}_{t},
    \mathbf{H}^{r}_{t}
    \right),
\end{equation}

where $\mathcal{O}_{\theta}$ is the learned temporal response operator. Here, $\mathbf{X}_{t}$ denotes the nodal coordinates, $\Delta\mathbf{U}_{t}$ is the current incremental displacement fields, $\mathbf{B}_{t}$ denotes the node-wise boundary-condition features, and $\mathbf{E}_{t}$ denotes the edge features in matrix form. Note that the inclusion of edge features is optional and depends on the chosen data representation. Scalar inputs and outputs are also included optionally when required by the prediction task. $\mathbf{H}^{r}_{t}$ is the recurrent mesh-state representation carried from the preceding prediction step. The recurrent state is internal to the model and is initialised as $\mathbf{H}^{r}_{0}=\mathbf{0}$ at the beginning of each simulation trajectory. Repeated application of $\mathcal{O}_{\theta}$ gives an autoregressive rollout over the crash event. During rollout, the predicted incremental displacement field and scalar-response vector are fed back as inputs for the next time step.

Given a trajectory length $T$, repeated application of the one-step operator produces the rollout prediction
\begin{equation}
    \left\{
    \Delta\widehat{\mathbf{U}}_{1:T},
    \widehat{\mathbf{s}}_{1:T}
    \right\}
    =
    \mathcal{R}_{\theta}
    \left(
    \mathbf{X}_{0},
    \Delta\mathbf{U}_{0},
    \mathbf{B}_{0},
    \mathbf{E}_{0},
    \mathbf{s}_{0}
    \right),
\end{equation}
where $\mathcal{R}_{\theta}$ denotes the autoregressive rollout operator. The accumulated displacement field is obtained by summing the predicted incremental displacement fields over time,
\begin{equation}
    \widehat{\mathbf{U}}_{t}
    =
    \sum_{k=1}^{t}
    \Delta\widehat{\mathbf{U}}_{k}.
\end{equation}
At each rollout step, the nodal coordinates and edge features are updated from the predicted displacement. The prescribed node types, distance encodings, support stiffnesses, and initial impactor velocity remain fixed throughout the rollout, while the temporal scalar-condition features are updated from the predicted scalar responses. This formulation allows the surrogate to predict both the transient deformation field and the scalar crashworthiness responses for unseen component geometries and imposed boundary conditions.

\section{Boundary-Condition-Aware Transformer Neural Operator}
\label{sec:method}

\begin{figure}[htbp]
    \centering
    \includegraphics[width=\textwidth]{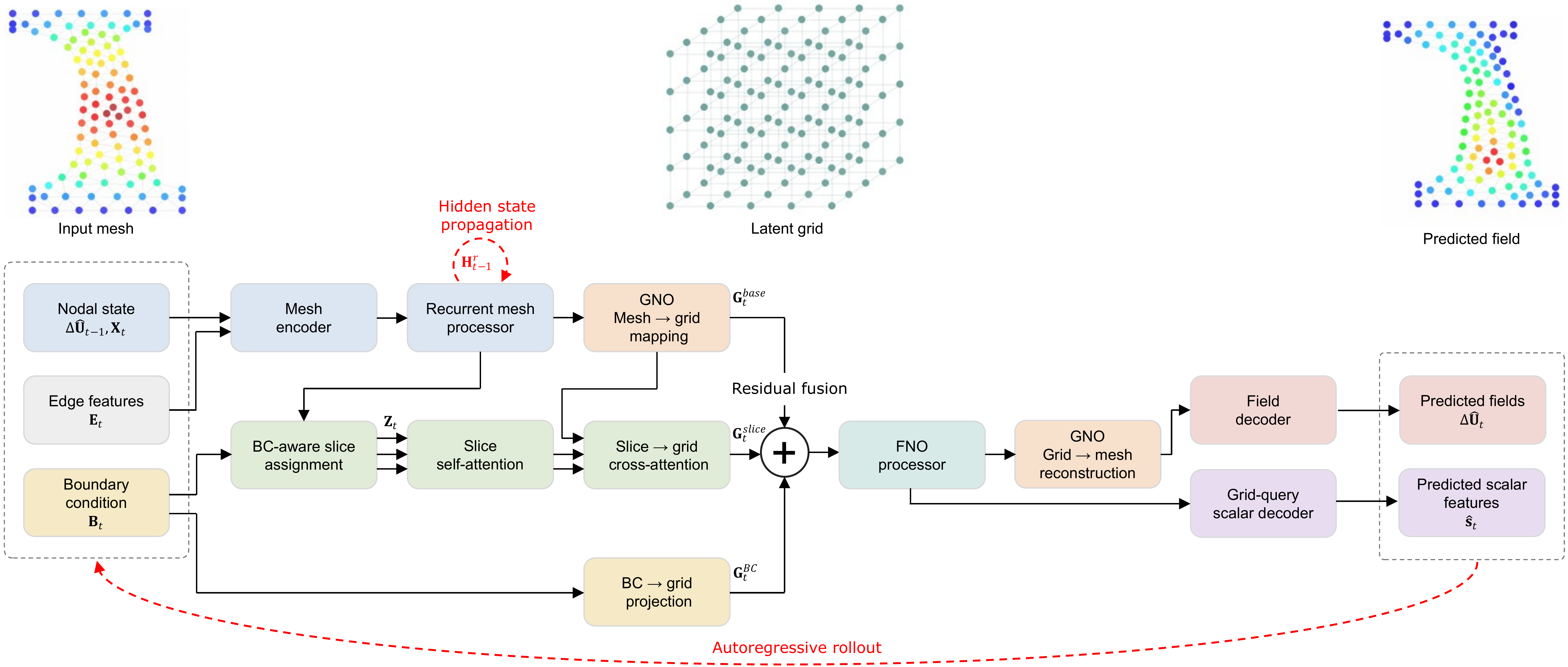}
    \caption{Overview of the proposed BAT-NO architecture.}
    \label{fig:bat_no_architecture}
\end{figure}

The proposed BAT-NO architecture is shown in Fig.~\ref{fig:bat_no_architecture}. The mesh encoder combines the nodal structural state with the node-wise condition features and encodes the edge features separately. A recurrent mesh processor then updates the resulting mesh representation while carrying temporal information between rollout steps. The processed mesh features are projected to a regular latent grid by a GNO layer. In parallel, the boundary-condition features are encoded by a slice-based transformer pathway and a direct boundary-to-grid projection. These three grid representations are fused and processed by FNO layers. The output grid features are finally reconstructed on the original mesh for field prediction, and queried by a scalar decoder for global crashworthiness responses.

\subsection{Mesh Input Decomposition and Encoding}
\label{subsec:mesh_encoding}

The model receives the nodal structural-state features, edge features, and node-wise condition features defined in Section~\ref{sec:problem_formulation}. The mesh encoder combines both node-level inputs $\mathbf{v}_{i}$ and $\mathbf{b}_{i}$, and maps them to the latent dimension,
\begin{equation}
    \mathbf{v}'_{i}
    =
    \phi_{\mathrm{v}}
    \left(
    \left[
    \mathbf{v}_{i},\,
    \mathbf{b}_{i}
    \right]
    \right),
    \qquad
    \mathbf{e}'_{ij}
    =
    \phi_{\mathrm{e}}
    \left(
    \mathbf{e}_{ij}
    \right),
\end{equation}
where $\phi_{\mathrm{v}}$ and $\phi_{\mathrm{e}}$ are MLP encoders. The encoded node feature $\mathbf{v}'_{i}$ is used as the initial hidden feature of the mesh processor. The same condition feature $\mathbf{b}_{i}$ is also passed directly to the slice-based and boundary-to-grid pathways. All input channels are normalised using statistics calculated from the training set.

\subsection{Recurrent Mesh Processor}
\label{subsec:recurrent_graph_processor}

The encoded node and edge features are processed by local message passing on the mesh graph. This transfers the relational and geometric information carried by the encoded edge features into the node representations. Let $\mathbf{h}_i^{(\ell)}$ be the hidden node feature at message-passing layer $\ell$, with $\mathbf{h}_i^{(0)}=\mathbf{v}'_i$. For an edge $(i,j)$, the edge message and its aggregation at node $i$ are computed as
\begin{equation}
    \mathbf{m}_{ij}^{(\ell)}
    = \phi_{\mathrm{m}}^{(\ell)}
    \left(\left[
    \mathbf{h}_i^{(\ell)},
    \mathbf{h}_j^{(\ell)},
    \mathbf{e}'_{ij}
    \right]\right),
    \qquad
    \mathbf{m}'_{i}{}^{(\ell)}
    = \sum_{j:(i,j)\in\mathcal{E}}
    \mathbf{m}_{ij}^{(\ell)}.
\end{equation}
The node feature is then updated through a residual MLP,
\begin{equation}
    \mathbf{h}_{i}^{(\ell+1)}
    =
    \mathbf{h}_{i}^{(\ell)}
    +
    \phi_{\mathrm{h}}^{(\ell)}
    \left(
    \left[
    \mathbf{h}_{i}^{(\ell)},
    \mathbf{m}'_{i}{}^{(\ell)}
    \right]
    \right).
\end{equation}

To include temporal information during autoregressive rollout, the final processed node feature is fused with the recurrent hidden state from the preceding time step. Let $\mathbf{h}^{(L_{\mathrm{MP}})}_{i,t}$ denote the output of the final mesh-processor layer at time step $t$, and let $\mathbf{h}^{r}_{i,t-1}$ denote the recurrent state carried from time step $t-1$. The recurrent state is initialised as $\mathbf{h}^{r}_{i,0}=\mathbf{0}$ for all $i\in\mathcal{V}$. The recurrent update is
\begin{equation}
    \boldsymbol{\beta}_{i,t}
    =
    \sigma
    \left(
    \phi_{\beta}
    \left(
    \left[
    \mathbf{h}^{(L_{\mathrm{MP}})}_{i,t},\,
    \mathbf{h}^{r}_{i,t-1},\,
    \mathbf{h}^{(L_{\mathrm{MP}})}_{i,t}
    -
    \mathbf{h}^{r}_{i,t-1}
    \right]
    \right)
    \right),
\end{equation}
\begin{equation}
    \widetilde{\mathbf{h}}^{r}_{i,t}
    =
    \phi_{\mathrm{r}}
    \left(
    \left[
    \mathbf{h}^{(L_{\mathrm{MP}})}_{i,t},\,
    \mathbf{h}^{r}_{i,t-1}
    \right]
    \right),
\end{equation}
\begin{equation}
    \mathbf{h}^{r}_{i,t}
    =
    \boldsymbol{\beta}_{i,t}
    \odot
    \widetilde{\mathbf{h}}^{r}_{i,t}
    +
    \left(
    1-\boldsymbol{\beta}_{i,t}
    \right)
    \odot
    \mathbf{h}^{r}_{i,t-1}.
\end{equation}
This is a simplified gated recurrence, where $\boldsymbol{\beta}_{i,t}$ controls the interpolation between the candidate hidden feature $\widetilde{\mathbf{h}}^{r}_{i,t}$ and the recurrent state $\mathbf{h}^{r}_{i,t-1}$ carried from the preceding time step. Here, $\sigma$ is the sigmoid function and $\odot$ denotes element-wise multiplication. The updated recurrent feature $\mathbf{h}^{r}_{i,t}$ is then passed to the mesh-to-grid and slice-based pathways.

\subsection{Mesh-to-Grid Mapping}
\label{subsec:mesh_to_grid}

Following the GINO formulation, we construct a regular 3D latent grid inside the minimum bounding box of the component geometry. The grid resolution remains fixed throughout the trajectory, while its physical coordinates are recomputed at each time step from the minimum bounding box of the current deformed mesh. During autoregressive rollout, the grid is therefore constructed from the predicted nodal coordinates at the corresponding time step. The grid nodes are denoted by
\begin{equation}
    \Xi =
    \left\{
    \boldsymbol{\xi}_q
    \right\}_{q=1}^{M},
\end{equation}
where $\boldsymbol{\xi}_q$ is the coordinate of grid node $q$ and $M$ is the number of grid nodes.

The recurrent mesh features are projected to the regular grid using a local GNO-style kernel projection. For grid node $q$, the base grid-feature vector is computed from neighbouring mesh nodes as
\begin{equation}
    \mathbf{g}^{\mathrm{base}}_q
    =
    \sum_{i\in\mathcal{N}(q)}
    w_{qi}
    \left[
    \sigma
    \left(
    \kappa_{\mathrm{down}}
    \left(
    \left[
    \boldsymbol{\xi}_q-\mathbf{x}_i,\,
    \left\|
    \boldsymbol{\xi}_q-\mathbf{x}_i
    \right\|_2
    \right]
    \right)
    \right)
    \odot
    \mathbf{h}^{r}_{i,t}
    \right],
\end{equation}
where $\boldsymbol{\xi}_q$ is the coordinate of grid node $q$, $\mathcal{N}(q)$ is the set of its $k$ nearest mesh nodes, $\kappa_{\mathrm{down}}$ is the learned mesh-to-grid kernel, $\sigma$ is the sigmoid function, and $\odot$ denotes element-wise multiplication. The scalar coefficient $w_{qi}$ is obtained by applying a distance softmax over the neighbouring mesh nodes,
\begin{equation}
    w_{qi}
    =
    \operatorname{softmax}_{i\in\mathcal{N}(q)}
    \left(
    -\frac{d_{qi}}{\eta}
    \right),
    \qquad
    d_{qi}
    =
    \left\|
    \boldsymbol{\xi}_q-\mathbf{x}_i
    \right\|_2 ,
\end{equation}
where $\eta$ controls the distance-weighting scale. In the implementation, $k=8$. For each sample and time step, $\eta$ is computed as the median of all $k$-nearest-neighbour distances associated with the current projection. It is a single scalar shared across all query points within that projection, rather than a pointwise quantity or a statistic calculated from the complete training set. The scale is recomputed separately for the mesh-to-grid and grid-to-mesh projections. The projected grid representation is then collected as
\begin{equation}
    \mathbf{G}^{\mathrm{base}}
    =
    \left[
    \mathbf{g}^{\mathrm{base}}_1,\ldots,
    \mathbf{g}^{\mathrm{base}}_M
    \right]^{\mathsf T}
    \in \mathbb{R}^{M\times d_h}.
\end{equation}
This operation gives a local kernel projection from the irregular mesh representation to the regular latent grid, following the mesh-to-grid mapping principle used in GINO. 

\subsection{Slice-Based Boundary-Condition Encoding}
\label{subsec:slice_encoding}

The slice-based transformer pathway uses the boundary-condition features to organise the processed mesh representation into a compact set of condition-aware slice tokens. For each node, the slice-assignment weights are computed from its boundary-condition features, and these weights determine how its hidden feature from the preceding mesh processor contributes to each slice token. The resulting tokens therefore contain the processed hidden node features grouped according to the boundary conditions. Multi-head self-attention models global interactions among the slice tokens, after which slice-to-grid cross-attention transfers the globally processed information to the latent grid. This pathway complements the local mesh-to-grid projection of the nodal state. 

First, each node is softly assigned to a set of latent slices according to its boundary-condition feature $\mathbf{b}_{i}$. For node $i$ and slice $k$, the assignment weight is computed as
\begin{equation}
    \alpha_{ik}
    =
    \operatorname{softmax}_{k}
    \left(
    \left[
    \phi_{\mathrm{s}}
    \left(
    \mathbf{b}_i
    \right)
    \right]_k
    \right),
\end{equation}
where $\mathbf{b}_{i}$ is the boundary-condition feature vector associated with node $i$, which will be described in detail in Section~\ref{subsec:input_representation}, and $\phi_{\mathrm{s}}$ maps this feature vector to the slice logits. The corresponding slice token is obtained by normalised aggregation of the recurrent mesh latent features,
\begin{equation}
    \mathbf{z}_{k}
    =
    \frac{
    \sum_{i\in\mathcal{V}}
    \alpha_{ik}
    \phi_{\mathrm{p}}
    \left(
    \mathbf{h}^{r}_{i,t}
    \right)
    }{
    \sum_{i\in\mathcal{V}}\alpha_{ik}+\epsilon
    },
\end{equation}
where $\phi_{\mathrm{p}}$ is a projection network and $\epsilon$ is a small constant for numerical stability. At transformer layer $\ell$, the slice tokens are collected as $\mathbf{Z}^{(\ell)}$.

At transformer layer $\ell$, the slice tokens $\mathbf{Z}^{(\ell)}$ are processed by a pre-normalised multi-head self-attention block with residual connections and a feed-forward network, producing the updated tokens $\mathbf{Z}^{(\ell+1)}$.

Both the slice self-attention and slice-to-grid cross-attention use standard multi-head scaled dot-product attention. Each attention head applies learned linear projections to the queries, keys, and values, and the outputs of all heads are concatenated and projected to the latent dimension. In the self-attention block, the queries, keys, and values are obtained from the slice tokens, allowing information exchange among the condition-aware slices. In the subsequent cross-attention block, the queries are obtained from the base-grid representation and grid coordinates, while the keys and values are obtained from the processed slice tokens. This asymmetric construction follows Perceiver-style cross-attention~\cite{jaegle2021perceiver,jaegle2021perceiverio} and transfers the globally processed mesh information from the condition-aware slices to the latent grid.
\begin{equation}
    \mathbf{Q}
    = \phi_{\mathrm{Q}}\left(\left[
    \mathbf{G}^{\mathrm{base}}, \boldsymbol{\Xi}
    \right]\right),
    \qquad
    \mathbf{K}
    = \phi_{\mathrm{K}}\left(\mathbf{Z}^{(\ell+1)}\right),
    \qquad
    \mathbf{V}
    = \phi_{\mathrm{V}}\left(\mathbf{Z}^{(\ell+1)}\right).
\end{equation}
The resulting slice-to-grid features are given by
\begin{equation}
    \mathbf{G}^{\mathrm{slice}}
    =
    \operatorname{CrossAttn}
    \left(
    \mathbf{Q},
    \mathbf{K},
    \mathbf{V}
    \right).
\end{equation}
This is later fused with the base grid representation $\mathbf{G}^{\mathrm{base}}$ using the trainable coefficient $\gamma_{\mathrm{slice}}$.

\subsection{Direct Boundary-Condition Projection and Residual Fusion}
\label{subsec:bc_projection_fusion}

The slice pathway provides a compact global representation of the processed mesh features organised by the boundary conditions. To retain the local spatial distribution of the boundary-condition features, they are also interpolated directly onto the latent grid using the distance-softmax weights from the mesh-to-grid projection. For grid node $q$, the directly conditioned grid feature is
\begin{equation}
    \mathbf{g}^{\mathrm{BC}}_q
    =
    \phi_{\mathrm{BC}}
    \left(
    \left[
    \mathbf{g}^{\mathrm{base}}_q,\,
    \sum_{i\in\mathcal{N}(q)}
    w_{qi}\mathbf{b}_{i}
    \right]
    \right)
    \in\mathbb{R}^{d_h},
\end{equation}
where $w_{qi}$ is the distance-softmax weight defined for the mesh-to-grid projection and $\phi_{\mathrm{BC}}:\mathbb{R}^{d_h+d_b}\rightarrow\mathbb{R}^{d_h}$ is an MLP.

The three grid representations are fused by residual addition,
\begin{equation}
    \mathbf{G}
    =
    \mathbf{G}^{\mathrm{base}}
    +
    \gamma_{\mathrm{slice}}
    \mathbf{G}^{\mathrm{slice}}
    +
    \gamma_{\mathrm{BC}}
    \mathbf{G}^{\mathrm{BC}},
\end{equation}
where $\gamma_{\mathrm{slice}}$ and $\gamma_{\mathrm{BC}}$ are trainable residual coefficients. They are initialised to $0.5$ and $1.0$, respectively, and optimised jointly with the remaining model parameters. The fused feature $\mathbf{G}$ is used as the input to the FNO processor.

\subsection{FNO Processing and Grid-to-Mesh Reconstruction}
\label{subsec:fno_reconstruction}

The fused grid representation is processed by a stack of FNO layers. Let $\mathbf{G}^{(0)}=\mathbf{G}$ denote the fused latent-grid feature. At each layer, the grid features are transformed into Fourier space and multiplied by learned complex-valued coefficients over a finite set of retained modes $\mathcal{I}\subset\mathbb{Z}^{3}$. The Fourier representation of the layer-specific kernel is denoted by $\mathcal{F}(\kappa_{\mathrm{FNO}}^{(\ell)})(\boldsymbol{\omega})$. Its coefficients are learned for $\boldsymbol{\omega}\in\mathcal{I}$ and set to zero for modes outside $\mathcal{I}$. The spectral convolution is written as
\begin{equation}
    \mathcal{K}^{(\ell)}
    \left(
    \mathbf{G}^{(\ell)}
    \right)
    =
    \mathcal{F}^{-1}
    \left(
    \mathcal{F}
    \left(
    \kappa_{\mathrm{FNO}}^{(\ell)}
    \right)
    \odot
    \mathcal{F}
    \left(
    \mathbf{G}^{(\ell)}
    \right)
    \right),
\end{equation}
where $\mathcal{F}$ and $\mathcal{F}^{-1}$ denote the Fourier transform and inverse Fourier transform on the regular grid, and $\odot$ denotes multiplication in Fourier space. The FNO layer update is
\begin{equation}
    \mathbf{G}^{(\ell+1)}
    =
    \sigma
    \left(
    \mathcal{W}^{(\ell)}
    \mathbf{G}^{(\ell)}
    +
    \mathcal{K}^{(\ell)}
    \left(
    \mathbf{G}^{(\ell)}
    \right)
    \right),
\end{equation}
where $\mathcal{W}^{(\ell)}$ is a point-wise linear transform and $\sigma$ denotes the nonlinear activation function.

The processed grid feature is reconstructed on the original mesh using a GNO-style grid-to-mesh kernel projection. Let $\mathbf{g}^{(L_{\mathrm{FNO}})}_q$ denote the output feature vector of the last FNO layer at grid node $q$. For mesh node $i$, the reconstructed mesh feature is
\begin{equation}
    \mathbf{h}^{o}_{i}
    =
    \sum_{q\in\mathcal{N}(i)}
    w_{iq}
    \left[
    \sigma
    \left(
    \kappa_{\mathrm{up}}
    \left(
    \left[
    \mathbf{x}_i-\boldsymbol{\xi}_q,\,
    \left\|
    \mathbf{x}_i-\boldsymbol{\xi}_q
    \right\|_2
    \right]
    \right)
    \right)
    \odot
    \mathbf{g}^{(L_{\mathrm{FNO}})}_q
    \right],
\end{equation}
where $\mathcal{N}(i)$ is the set of the $k$ nearest grid nodes to mesh node $i$, $\kappa_{\mathrm{up}}$ is the learned grid-to-mesh reconstruction kernel, and $\sigma$ is the sigmoid function. Analogously to the mesh-to-grid mapping, the scalar coefficient $w_{iq}$ is obtained by applying a softmax to the negative Euclidean distances between mesh node $i$ and its neighbouring grid nodes. The same number of neighbours, $k=8$, is used for both projections. The distance scale $\eta$ follows the same definition but is recomputed for the grid-to-mesh projection as the median of its corresponding $k$-nearest-neighbour distances.

\subsection{Field and Scalar Decoders}
\label{subsec:decoders}

The reconstructed mesh feature is decoded to the next incremental displacement by an MLP,
\begin{equation}
    \Delta\widehat{\mathbf{u}}_{i}
    =
    \phi_{\mathrm{field}}
    \left(
    \mathbf{h}^{o}_{i}
    \right).
\end{equation}
During autoregressive rollout, the predicted incremental displacement is accumulated to update the deformed nodal coordinates for the next step.

For scalar-response prediction, BAT-NO uses a cross-attention-based decoder. Let $\mathbf{Q}_{s}\in\mathbb{R}^{d_s\times d_h}$ denote the learned scalar-query tokens, where $d_s$ is the number of scalar responses and $d_h=128$ is the latent dimension. For the grid-query decoder, the keys $\mathbf{K}_{s}$ and values $\mathbf{V}_{s}$ are calculated from the processed grid representation $\mathbf{G}^{(L_{\mathrm{FNO}})}$ using separate learned linear layers. The scalar-query tokens then attend to these grid-derived features,
\begin{equation}
    \mathbf{S}'
    =
    \mathbf{Q}_{s}
    +
    \operatorname{CrossAttn}
    \left(
        \operatorname{LayerNorm}(\mathbf{Q}_{s}),
        \operatorname{LayerNorm}(\mathbf{K}_{s}),
        \operatorname{LayerNorm}(\mathbf{V}_{s})
    \right),
\end{equation}
where $\operatorname{CrossAttn}$ denotes multi-head cross-attention. The scalar tokens are further updated by a feed-forward network, after which the scalar-response vector is predicted by scalar output heads,
\begin{equation}
\begin{aligned}
    \mathbf{S}
    &=
    \mathbf{S}'
    +
    \operatorname{FFN}
    \left(
        \operatorname{LayerNorm}(\mathbf{S}')
    \right),
    \\
    \widehat{\mathbf{s}}
    &=
    \phi_{\mathrm{scalar}}
    \left(
    \mathbf{S}
    \right).
\end{aligned}
\end{equation}
In the default BAT-NO variant, the scalar-query tokens attend to the processed grid features. We also consider a mesh-query variant, where the same decoder is applied after grid-to-mesh reconstruction by replacing $\mathbf{G}^{(L_{\mathrm{FNO}})}$ with the reconstructed mesh features $\mathbf{H}^{o}=\{\mathbf{h}^{o}_i\}_{i\in\mathcal{V}}$. The grid-query decoder extracts scalar information from the globally processed latent-grid representation, while the mesh-query decoder attends to the reconstructed node-level features and therefore retains more local spatial detail. The grid representation may be well suited to global responses such as internal energy, whereas the reconstructed mesh representation may benefit localised responses such as support reaction forces. During autoregressive rollout, the predicted scalar-response vector is fed back together with the predicted displacement field, without substituting ground-truth responses at subsequent time steps.

\section{Datasets and Experimental Setup}
\label{sec:datasets_experiments}

\subsection{Datasets}
\label{subsec:datasets}

This subsection describes the datasets used for training and evaluating the model. The simulations use the component-level B-pillar side-impact FE model introduced in~\cite{li2026regunet}. The scaled hot-stamped boron-steel B-pillar is discretised using triangular shell elements with a uniform thickness of $1.6$~mm and an average element size of approximately $3$~mm, resulting in approximately $4{,}000$ nodes and $7{,}000$--$8{,}000$ shell elements. A strain-rate-dependent material description of fully martensitic boron steel is used. Impact is applied by a rigid impactor with a diameter of $40$~mm and a mass of $100$~kg. Surface-to-surface contact is defined between the impactor and the component using a friction coefficient of $0.1$. The simulations are performed using an explicit time-integration scheme in Virtual Performance Solution. The FE implementation used in this study, including the fixed and elastic support definitions, is summarised in Appendix~\ref{app:fe_implementation}.

Each simulation covers a total duration of $30$~ms. The displacement fields and scalar responses are extracted at intervals of $1.5$~ms. The first 12 response increments are retained for model training and evaluation, covering the period in which the principal B-pillar deformation develops. Each simulation is therefore represented as a temporal graph sample containing 12 consecutive prediction steps.

\begin{figure}[htbp]
    \centering
    \includegraphics[width=0.7\linewidth]{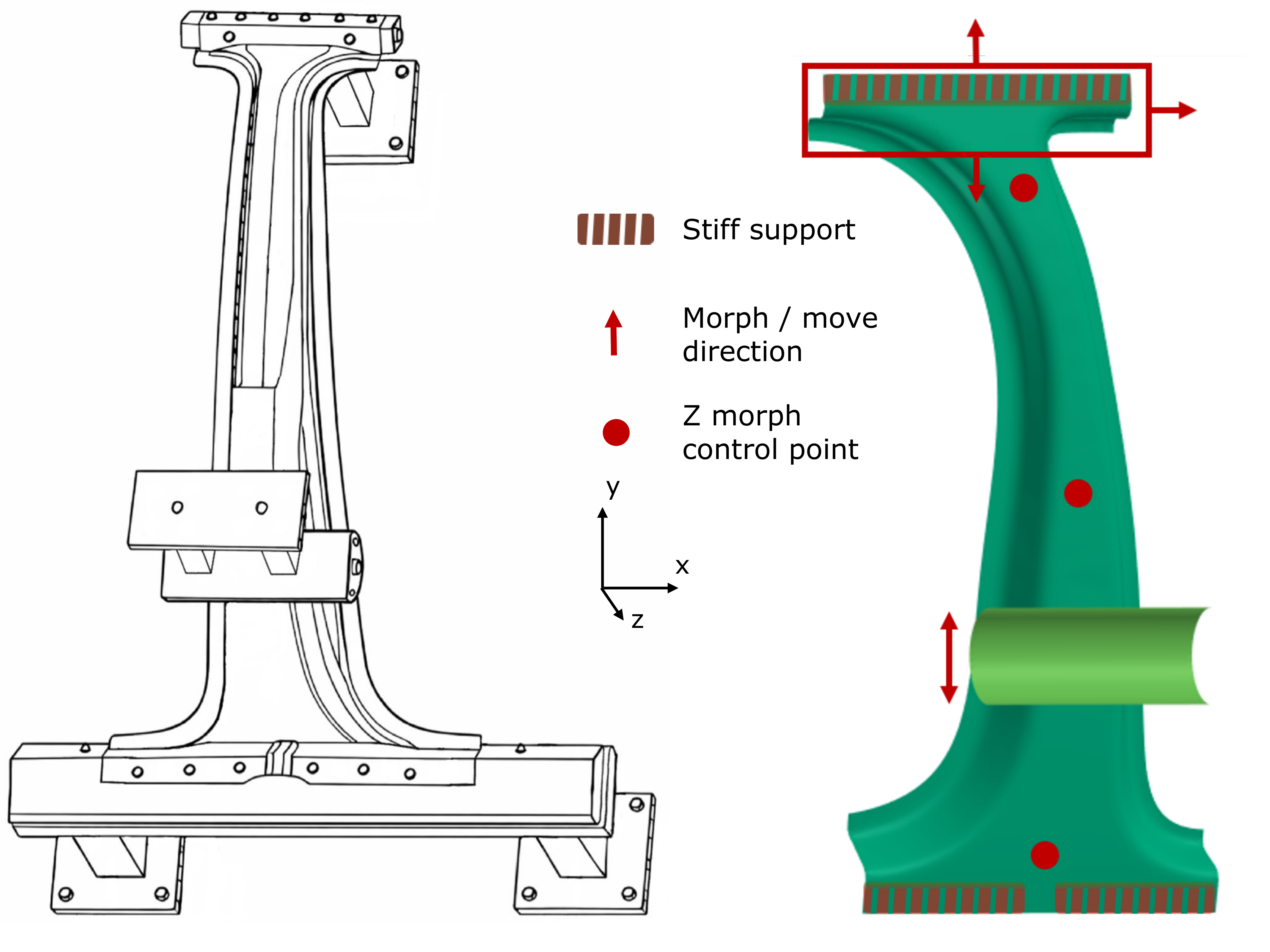}
    \caption{FE side-impact setup and geometric morphing scheme used to generate Cases S, SL, and SLB. The illustration shows the B-pillar, impactor, connection regions, and the geometric control parameters used for shape variation.}
    \label{fig:dataset-fe-morphing-setup}
\end{figure}

Three datasets are generated from component-level side-impact simulations of a B-pillar component of a vehicle. The FE setup, loading configuration, support locations, and geometric morphing scheme are illustrated in Fig.~\ref{fig:dataset-fe-morphing-setup}. The datasets progressively expand the design space from shape variation alone to simultaneous variations in shape, loading, and boundary conditions. These cases provide progressively broader prediction settings for evaluating model performance as loading and boundary-condition variations are introduced alongside geometry variation.

The first dataset, denoted \textbf{Case S}, contains shape variation only. The B-pillar geometry is modified through mesh morphing while the impactor remains at its reference impact position of $90$ mm with an initial velocity of $4$~m/s, and the upper and lower component regions remain fixed in all six degrees of freedom. The upper region of the component is morphed in the $x$ and $y$ directions, and the $z$-direction morph is applied through one of three predefined control points on the component. These operations generate B-pillars with distinct shapes, which isolates the effect of component geometry on the crash response.

The second dataset, denoted \textbf{Case SL}, introduces loading variation in addition to shape variation. The impactor is translated along the $y$ direction to vary the vertical impact position, thereby simulating component-level impact tests conducted at different vehicle suspension heights. Also, the impactor's initial velocity is varied to change the impact severity. The boundary conditions remain fixed in all six degrees of freedom. The loading parameters cover impact positions from 75 to 105~mm and initial velocities from 3.5 to 4.0~m/s. The morphing ranges used in Case SL are narrower than those used in Case S, in order to compensate for the two additional loading variables and maintain a comparable overall level of normalised parameter variation under a fixed simulation budget. The detailed design-space justification is provided in \ref{app:dataset_design_space}.

The third dataset, denoted \textbf{Case SLB}, jointly varies the shape, loading, and boundary conditions. It uses the wider shape ranges of Case S and the loading ranges of Case SL. The fixed boundary constraints are replaced by elastic supports at the top, bottom-left, and bottom-right regions. Each support stiffness is sampled independently between 1 and 10~kN/mm, allowing asymmetric constraint conditions and different support reaction forces during impact. These elastic supports provide a simplified representation of the B-pillar connections to neighbouring components in an assembly. Changes to adjacent components alter the response transmitted through these connections. Independently varying the three stiffnesses approximates this assembly-level influence within the component model. The complete parameter ranges for the three cases are summarised in Table~\ref{tab:dataset-parameter-ranges} in \ref{app:dataset_design_space}.

The geometry, loading, and support parameters are sampled using Latin hypercube sampling (LHS) over the ranges defined for each case. Independent LHS designs are used to generate 100 training simulations and 50 validation simulations for each case. The validation sets are used for model comparison, ablation, and hyperparameter selection. For the final evaluation of Case SLB, a separate test set of 300 simulations is generated independently over the same parameter ranges. This test set represents held-out in-design-space samples and is not used for model training or hyperparameter selection. All input and output normalisation statistics are calculated from the corresponding training set and then fixed for validation and test evaluation.

\subsection{Input Representation}
\label{subsec:input_representation}

At each time step, the model input is constructed from the current mesh state and the condition information defined on the component. The dynamic state is represented by the nodal incremental displacement $\Delta\mathbf{u}_{i,t}$ defined in Eq.~\eqref{eq:incremental-displacement}. The model predicts the next incremental displacement $\Delta\widehat{\mathbf{u}}_{i,t+1}$, which is accumulated during rollout to recover the transient deformation history.

Boundary and contact information is encoded as node-wise condition features. The node type is represented by a one-hot vector $\mathbf{c}_i$ that distinguishes free, constraint, and contact nodes. For cases with elastic supports, the constraint nodes are further divided into top, bottom-left, and bottom-right constraint types. For the fixed-boundary cases, all constraint nodes are represented by a single constraint-node type. To provide spatial information about the location of critical regions, each node is also assigned distance-based positional encodings. These encodings are computed from the Euclidean distance to the nearest critical nodes, where the critical-node sets correspond to the constraint and contact regions,
\begin{equation}
    d_i^{\mathcal{C}}
    =
    \min_{j\in\mathcal{C}}
    \left\|
    \mathbf{x}_i-\mathbf{x}_j
    \right\|_2 .
\end{equation}
Here, $\mathcal{C}$ denotes a critical-node set and $\mathbf{x}_i$ is the nodal coordinate. The critical-node sets correspond to the constraint and contact regions prescribed in the FE setup. These sets are defined for each sample before the rollout and are kept fixed throughout the trajectory.

The static boundary and loading parameters comprise the top-support stiffness, bottom-left-support stiffness, bottom-right-support stiffness, and initial impactor velocity. Each support stiffness is assigned to its corresponding constraint nodes, while the initial impactor velocity is assigned to the contact nodes. This sparse assignment keeps the condition field spatially local rather than repeating all scalar parameters over the whole mesh. It informs the model where each parameter acts directly and which regions of the component it is expected to affect most strongly.

For simulations with scalar-response prediction, the temporal scalar features comprise internal energy, top spring force, bottom-left spring force, bottom-right spring force, contact force, and impactor intrusion. The scalar input at time step $t$ is taken from the previous step, and the initial scalar input is set to zero. Internal energy is assigned to all nodes, each spring force is assigned to its corresponding constraint nodes, and the contact force and impactor intrusion are assigned to the contact nodes.

The node-wise condition feature is obtained by concatenating the node-type indicators, distance-based encodings, static condition parameters, and temporal scalar features,
\begin{equation}
    \mathbf{b}_{i,t}
    =
    \left[
    \mathbf{c}_{i},\,
    \mathbf{d}_{i},\,
    \mathbf{p}^{\mathrm{static}}_{i},\,
    \mathbf{p}^{\mathrm{scalar}}_{i,t}
    \right],
\end{equation}
where $\mathbf{c}_{i}$ contains the node type one-hot vector, $\mathbf{d}_{i}$ contains the distance-based encodings, $\mathbf{p}^{\mathrm{static}}_{i}$ contains the static boundary and loading parameters, and $\mathbf{p}^{\mathrm{scalar}}_{i,t}$ contains the temporal scalar inputs. Together, $\mathbf{v}_{i,t}$ and $\mathbf{b}_{i,t}$ provide the complete node-level input to the mesh encoder.

\subsection{Training and Evaluation Protocol}
\label{subsec:training_evaluation}

The model is trained by minimising the mean squared error (MSE) between the predicted and target incremental displacement fields. For a trajectory with $T$ time steps, the field loss is
\begin{equation}
    \mathcal{L}_{\mathrm{field}}
    =
    \frac{1}{T|\mathcal{V}|}
    \sum_{t=1}^{T}
    \sum_{i\in\mathcal{V}}
    \left\|
    \Delta\widehat{\mathbf{u}}_{i,t}
    -
    \Delta\mathbf{u}_{i,t}
    \right\|_2^2 .
\end{equation}
When scalar crashworthiness responses are predicted, an additional scalar loss is included,
\begin{equation}
    \mathcal{L}_{\mathrm{scalar}}
    =
    \frac{1}{T}
    \sum_{t=1}^{T}
    \left\|
    \widehat{\mathbf{s}}_{t}
    -
    \mathbf{s}_{t}
    \right\|_2^2 .
\end{equation}
The total loss is
\begin{equation}
    \mathcal{L}
    =
    \mathcal{L}_{\mathrm{field}}
    +
    \lambda_{\mathrm{scalar}}
    \mathcal{L}_{\mathrm{scalar}},
\end{equation}
where $\lambda_{\mathrm{scalar}}$ controls the relative weight of the scalar-response prediction.

Evaluation is performed using a fully autoregressive rollout. At the first time step, the model receives the initial mesh state, prescribed condition features, and a zero-valued scalar-response history. At each subsequent step, the predicted incremental displacement is used to update the nodal displacement state and coordinates, from which the deformed edge attributes are recalculated. The initially zero-valued recurrent hidden state is propagated between consecutive time steps. When scalar prediction is enabled, the predicted scalar-response vector, including the predicted impactor intrusion, is sparsely reassigned to the corresponding node regions and used as the scalar input for the next step. No ground-truth displacement or scalar response from future time steps is used during rollout evaluation.

The main field metric is the final-step mean nodal Euclidean displacement error (MED),
\begin{equation}
    \varepsilon_{\mathrm{MED}}
    =
    \frac{1}{|\mathcal{V}|}
    \sum_{i\in\mathcal{V}}
    \left\|
    \widehat{\mathbf{u}}_{i,T}
    -
    \mathbf{u}_{i,T}
    \right\|_2 ,
\end{equation}
where $\widehat{\mathbf{u}}_{i,T}$ and $\mathbf{u}_{i,T}$ are the predicted and reference accumulated-displacement vectors at the final time step. The Euclidean displacement error is calculated at each node and then averaged over the mesh. MED is reported in millimetres and directly measures the average nodal error in the final deformed component shape.

We also report the MED at each rollout step to examine how the accumulated-displacement error develops during autoregressive prediction. For scalar response $m$ of sample $n$, the relative trajectory error is defined over the complete rollout as
\begin{equation}
    \varepsilon^{\mathrm{traj}}_{n,m}
    =
    100
    \frac{
    \left[
    \sum_{t=1}^{T}
    \left(
    \widehat{s}_{n,m,t}-s_{n,m,t}
    \right)^2
    \right]^{1/2}
    }{
    \max\left(
    \left[
    \sum_{t=1}^{T}
    s_{n,m,t}^{2}
    \right]^{1/2},
    \epsilon
    \right)
    },
\end{equation}
where $\epsilon$ is a small positive constant preventing division by zero. The reported error for response $m$ is the mean of $\varepsilon^{\mathrm{traj}}_{n,m}$ over the evaluated samples. The overall scalar error is obtained by averaging the six response-specific mean errors.

For each derived crashworthiness indicator $\mathcal{M}_p$, the absolute relative error is
\begin{equation}
    \varepsilon^{\mathrm{phys}}_{n,p}
    =
    100
    \frac{
    \left|
    \widehat{\mathcal{M}}_{n,p}
    -
    \mathcal{M}_{n,p}
    \right|
    }{
    \max\left(
    \left|\mathcal{M}_{n,p}\right|,
    \epsilon
    \right)
    }.
\end{equation}
The derived indicators comprise the peak support and contact forces, their time-integrated impulses, the final internal energy, and the maximum absolute impactor intrusion. The physical-response tables report the median error across the evaluated samples.

All models are trained with the same optimiser and loss functions within each case study. We also evaluate temporal training using both teacher forcing and scheduled sampling to assess the effect of predicted-state feedback on autoregressive rollout. This comparison, together with the learning rate, batch size, scalar-loss coefficient, and other detailed metric definitions, is reported in \ref{app:hyperparameter_tuning}. The final training configuration is summarised in Table~\ref{tab:training_details}.

\section{Results}
\label{sec:results}

\subsection{Baseline comparison}

This subsection evaluates BAT-NO by comparing its field-prediction accuracy with representative state-of-the-art graph-based, neural-operator, and transformer-based surrogate models. Table~\ref{tab:validation_accumulated_l2} compares BAT-NO with MGN~\cite{pfaff2021meshgraphnets}, ReGUNet~\cite{li2026regunet}, DeepONet~\cite{lu2021deeponet}, FNO~\cite{li2021fno}, GNO~\cite{li2020gno}, GINO~\cite{li2023gino}, Transolver~\cite{wu2024transolver}, and UPT~\cite{alkin2024upt}. Each model is trained and evaluated separately on Cases S, SL, and SLB using the same training and validation samples within each case. This comparison considers field prediction only and does not include scalar decoders or scalar-response feedback. The baseline hyperparameters are not separately tuned, each model uses the adopted reference configuration. BAT-NO is likewise evaluated using its untuned configuration, without model-specific hyperparameter optimisation. All reported values in Table~\ref{tab:validation_accumulated_l2} are mean errors over the 50 validation simulations obtained from fully autoregressive rollouts of one trained instance of each model. The corresponding sample-wise training and validation error distributions are presented in Appendix~\ref{app:train_validation_boxplots}. BAT-NO achieves the lowest mean validation error in all three case studies. Relative to the second-best model, it reduces the mean error by 18.2\% in Case S, from $0.1611$ to $0.1318$~mm. BAT-NO also achieves the best performance in Case SL, demonstrating consistent accuracy when the geometry and loading conditions vary while the boundary conditions remain fixed.

\begin{table}[htbp]
\centering
\caption{Mean final-step MED over the 50 validation samples for the three case studies. Errors are reported in millimetres. The best result in each case is shown in bold, and the second-best result is underlined.}
\label{tab:validation_accumulated_l2}
\begin{tabular}{lccc}
\toprule
Model & Case S & Case SL & Case SLB \\
\midrule
MGN~\cite{pfaff2021meshgraphnets} & 1.8967 & 0.5925 & 2.1354 \\
ReGUNet~\cite{li2026regunet} & \underline{0.1611} & 0.4356 & 1.1057 \\
DeepONet~\cite{lu2021deeponet} & 0.3239 & 0.3690 & 0.7607 \\
FNO~\cite{li2021fno} & 0.2181 & 0.2873 & 0.7772 \\
GNO~\cite{li2020gno} & 0.5555 & 0.3952 & 1.6343 \\
GINO~\cite{li2023gino} & 0.3796 & \underline{0.1752} & \underline{0.6689} \\
Transolver~\cite{wu2024transolver} & 0.3926 & 0.2981 & 0.7422 \\
UPT~\cite{alkin2024upt} & 0.3189 & 0.3416 & 0.6891 \\
\midrule
\textbf{BAT-NO (Ours)} & \textbf{0.1318} & \textbf{0.1742} & \textbf{0.4510} \\
\bottomrule
\end{tabular}
\end{table}

Several models, including MGN, GINO, GNO, and Transolver, exhibit lower errors in Case SL than in Case S, despite the introduction of loading variation. Case SL uses narrower geometry ranges than Case S, which may offset the additional complexity introduced by the varying loading condition. The observed differences therefore reflect the combined changes in geometry and loading ranges between the two benchmark spaces.

For BAT-NO, the mean error increases from $0.1318$~mm in Case S to $0.1742$~mm in Case SL. Case SL uses a narrower range of shape variation than Case S, while additionally varying the impact position and initial velocity. The increase in error occurs despite the narrower geometric design space and is consistent with the added loading variation introducing further prediction complexity. BAT-NO achieves the lowest mean error in Case SL, although its performance is comparable to GINO and the difference between the two models is small. This result suggests that both latent-grid operator architectures can incorporate the loading variation effectively, while BAT-NO maintains a slight advantage in mean field-prediction accuracy.

The advantage of BAT-NO becomes more substantial in Case SLB, where geometry, loading, and boundary conditions vary simultaneously. BAT-NO achieves a mean error of $0.4510$~mm, compared with $0.6689$~mm for the second-best model, GINO, corresponding to a relative error reduction of 32.6\%. The larger improvement suggests that the complete BAT-NO architecture is particularly effective in this setting. Its boundary-condition-guided slice representation, direct boundary-to-grid projection, recurrent mesh processing, and latent-grid operator learning jointly capture the local and global effects of varying geometry, loading, and boundary conditions.

\begin{figure}[htbp]
    \centering
    \includegraphics[width=\textwidth]{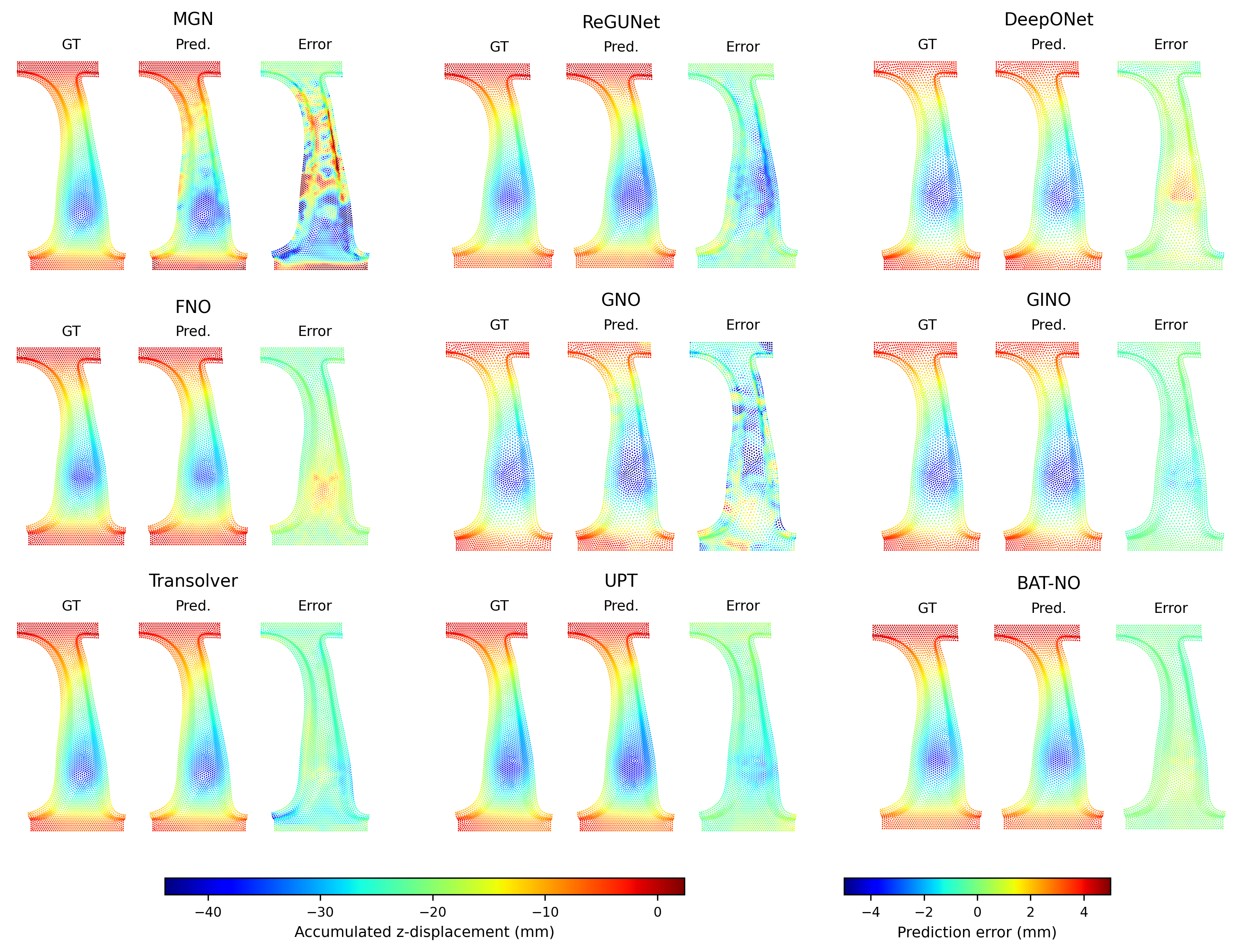}
    \caption{Qualitative baseline comparison of the accumulated $z$-displacement field. For each model, the panels show the ground-truth field (GT), predicted field (Pred.), and signed prediction-error field for that model's median-error validation sample. Displacement and error values are reported in millimetres.}
    \label{fig:baseline_visualisation}
\end{figure}

Figure~\ref{fig:baseline_visualisation} presents the displacement-field predictions for Case SLB. For each model, the visualised case is the validation sample whose final-step accumulated-displacement $L_2$ error is closest to the median of that model's validation-error distribution. The figure therefore illustrates the typical qualitative prediction produced by each model, rather than a comparison on a shared sample. The BAT-NO prediction shows close agreement with its corresponding ground-truth field and produces relatively small and smooth error patterns, consistent with the quantitative error distribution reported in Table~\ref{tab:validation_accumulated_l2}.

\subsection{Ablation Study}

The contribution of each BAT-NO component is examined through the incremental ablation presented in Table~\ref{tab:batno_component_ablation}. Each variant adds one component to the preceding configuration. The differences between adjacent variants therefore quantify the incremental contribution of each component. The base model follows the GINO encoder--operator--decoder architecture, comprising a GNO-based mesh-to-grid encoder, FNO processing on the latent grid, and a GNO-based grid-to-mesh decoder. It does not include pre-grid mesh message passing, boundary-condition-aware slice attention, temporal recurrence, or direct boundary-condition-to-grid projection.

\begin{table}[htbp]
\centering
\caption{Mean final-step MED over the 50 Case SLB validation simulations for the BAT-NO component ablation. Errors are reported in millimetres.}
\label{tab:batno_component_ablation}
\begin{tabular}{lc}
\toprule
Variant & Mean validation error (mm) \\
\midrule
Base (GINO) & 0.6689 \\
Base + Slice & 0.6139 \\
Base + Slice + MP & 0.5774 \\
Base + Slice + MP + Rec. & 0.5282 \\
\textbf{BAT-NO} & \textbf{0.4510} \\
\bottomrule
\end{tabular}
\end{table}

\begin{figure}[htbp]
    \centering
    \includegraphics[width=0.85\textwidth]{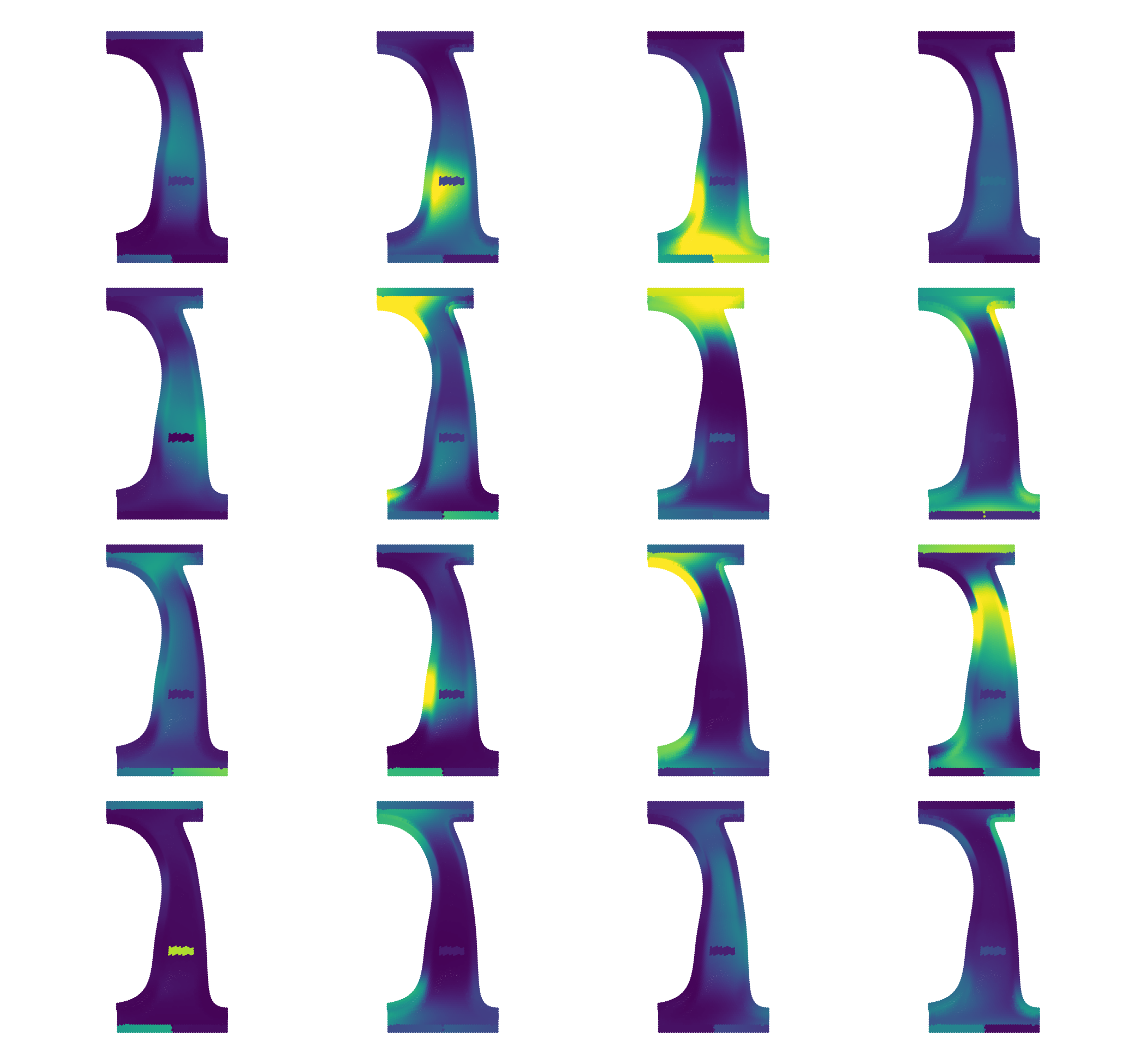}
    \caption{Visualisation of the learned node-to-slice assignments for a 16-slice example. Each panel corresponds to one slice, and the colour indicates the assignment score, representing how strongly each node is assigned to that slice.}
    \label{fig:slice_visualisation}
\end{figure}

\begin{figure}[htbp]
    \centering
    \includegraphics[width=0.98\textwidth]{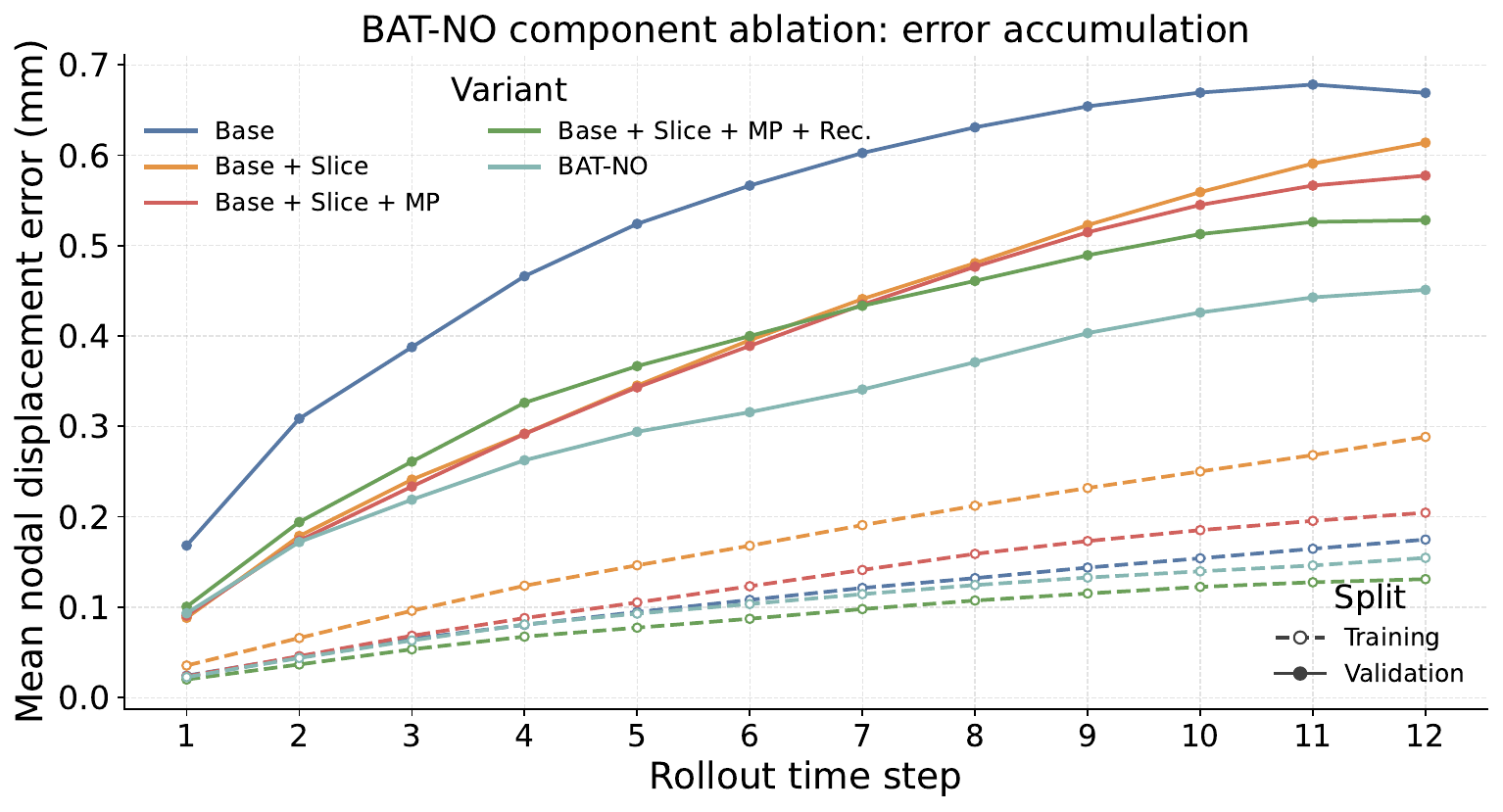}
    \caption{Mean nodal Euclidean displacement error over the autoregressive rollout for the BAT-NO component ablation on Case SLB. Dashed and solid lines denote the training and validation splits, respectively.}
    \label{fig:ablation_rollout}
\end{figure}

Introducing the boundary-condition-aware slice pathway reduces the mean validation error from $0.6689$ to $0.6139$~mm, corresponding to an improvement of 8.2\%. The training error increases while the validation error decreases, resulting in a smaller train--validation gap. This suggests that the slice pathway improves generalisation by organising the boundary-condition features according to learned physical relationships. Figure~\ref{fig:slice_visualisation} shows that the model distinguishes physically related nodes and assigns them high scores within the same slices. These learned groupings provide a more physically meaningful representation of the boundary conditions than the local nodal features alone. Self-attention captures interactions among the slices, while the subsequent slice-to-grid cross-attention transfers this global boundary-condition context to each grid location. The rollout curves in Figure~\ref{fig:ablation_rollout} show that the slice pathway reduces the mean validation error at every time step during the entire crash simulation.

\begin{figure}[htbp]
    \centering
    \includegraphics[width=0.98\textwidth]{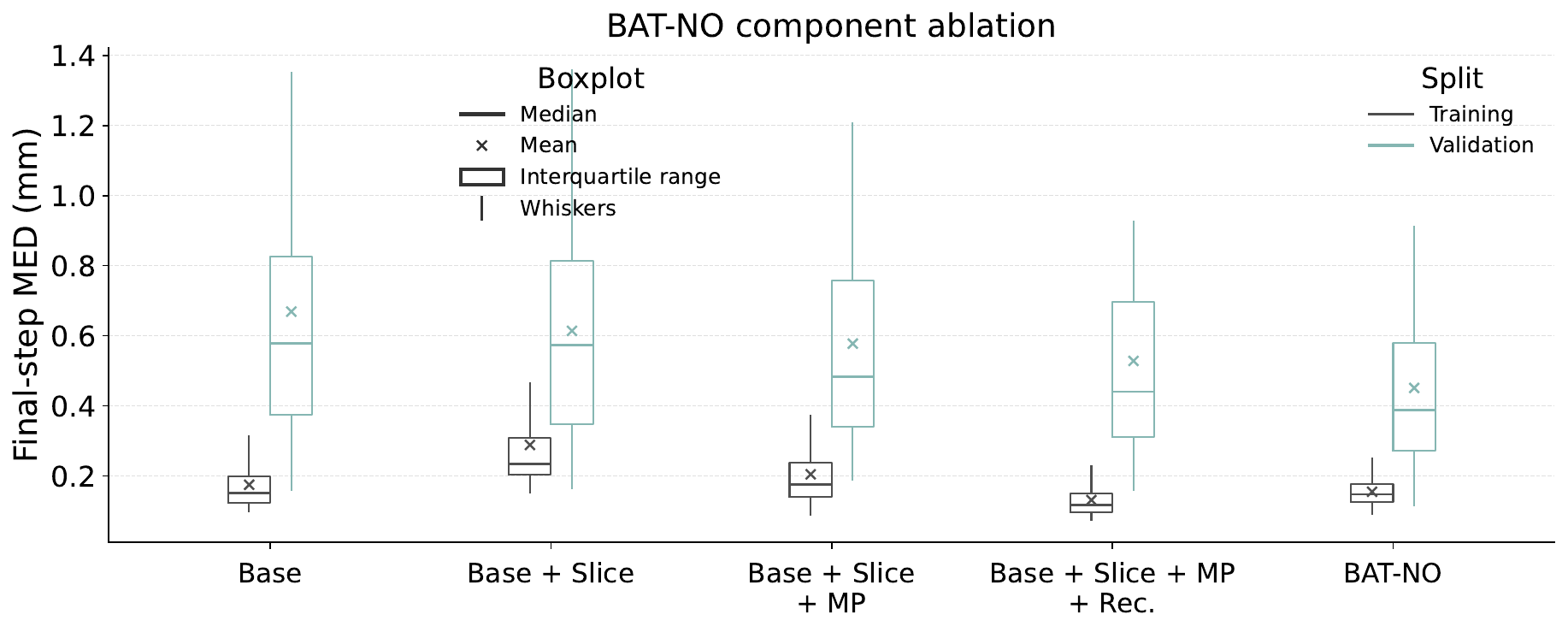}
    \caption{Training and validation distributions of the final-step MED for the BAT-NO component ablation on Case SLB. The box plots show how prediction errors change as the slice mapping, pre-projection message passing (MP), recurrence block (Rec.), and complete BAT-NO architecture are introduced.}
    \label{fig:ablation_boxplot}
\end{figure}

Adding the pre-grid mesh-processing block further reduces the mean validation error from $0.6139$ to $0.5774$~mm, giving an additional improvement of 5.9\%. Without this block, the features passed to the slice and mesh-to-grid pathways are obtained directly from the nodal encoder. The message-passing block incorporates the fine-mesh connectivity and edge features into the nodal representation before projection. Consequently, the information transferred to the latent grid contains both nodal state information and local geometric relationships.

Temporal recurrence reduces the mean error by a further 8.5\%, from $0.5774$ to $0.5282$~mm. Its effect is most evident in the rollout curves. The recurrent variant produces slightly higher validation errors during the early time steps, but its error becomes lower than that of the non-recurrent variant during the later rollout. The final-step error is consequently reduced. This behaviour indicates that the recurrent state preserves information from previous increments and helps suppress error accumulation over the longer rollout.

Finally, direct boundary-condition-to-grid projection produces the complete BAT-NO architecture. This component reduces the mean error from $0.5282$ to $0.4510$~mm, representing a further incremental improvement of 14.6\%. Relative to the base model, the complete architecture reduces the mean error by 32.6\%. It also achieves the lowest mean validation error throughout the rollout. The slice pathway provides a global, condition-aware representation by aggregating information across the mesh. In contrast, the direct boundary-condition projection transfers spatially local condition features to nearby grid locations. Combining these complementary pathways allows the latent grid to retain both the global configuration and local spatial distribution of the boundary conditions, improving its ability to represent changes in the critical regions and predict the resulting displacement field.

\subsection{Scalar Prediction Accuracy}

This experiment extends the untuned Case SLB BAT-NO configuration with joint field and scalar prediction. Three scalar-decoder variants are trained and compared: global pooling, grid query, and mesh query. Each variant predicts the incremental-displacement field and six scalar-response trajectories. During validation rollout, both the predicted field and predicted scalar responses are fed back at subsequent time steps. The comparison therefore evaluates decoder choice under the same fully coupled autoregressive protocol. The global-pooling decoder applies an MLP to the concatenated mean- and max-pooled features of the latent grid nodes, providing a direct and simple benchmark. The grid-query decoder uses cross-attention between trainable scalar-query tokens and the latent grid-node features, followed by an MLP decoder, as described in Section~\ref{subsec:decoders}. The mesh-query decoder uses the same attention-based formulation, but performs cross-attention between the scalar-query tokens and the reconstructed mesh features after the GNO grid-to-mesh layer.

Table~\ref{tab:scalar_trajectory_error} shows that both attention-based decoders outperform the global-pooling benchmark in mean scalar-response accuracy. The mesh-query decoder provides the lowest errors for the three spring reaction forces. These responses are spatially localised because each force acts only at its corresponding support region, whose location remains fixed across all samples. By querying the full reconstructed mesh, the mesh-query decoder retains direct information about the nodes associated with each support. In contrast, internal energy is a global response and is assigned to all nodes during preprocessing, for which the grid-query decoder provides higher accuracy. Contact force and impactor intrusion are also localised responses. Their prescribed contact regions vary across samples according to the impactor position but remain fixed during each rollout trajectory. The mesh-query decoder must therefore identify the relevant sample-dependent contact region within the full reconstructed mesh, whereas the grid-query decoder provides slightly lower errors for these quantities. Although the mesh-query decoder achieves a slightly lower mean scalar error, it produces a higher field-prediction error in the decoder comparison. Under the evaluated $8^3$ configuration, it also requires 53.0~s per epoch and 5.322~GB of peak GPU memory, compared with 48.2~s and 5.180~GB for the grid-query decoder. Together with its lower errors for most derived physical-response indicators, these results motivate the selection of the grid-query decoder for the final BAT-NO model. The measured cost comparison applies specifically to the $8^3$ decoder experiment.

Table~\ref{tab:physical_response_error} evaluates critical crashworthiness indicators derived from the predicted scalar-response trajectories. These indicators include the peak support and contact forces, the corresponding force impulses, the final internal energy, and the maximum impactor intrusion. The grid-query decoder achieves the lowest error for most of these indicators, demonstrating that it provides the best overall balance between scalar-response accuracy, displacement-field accuracy, and computational cost.

\begin{table}[htbp]
    \centering
    \caption{Mean relative trajectory errors for scalar-output prediction on the validation set and computational cost during training. Error values are percentages; lower is better. The best value in each row is shown in bold, and the second-best value is underlined.}
    \label{tab:scalar_trajectory_error}
    \small
    \begin{tabularx}{\linewidth}{@{}Xccc@{}}
        \toprule
        Scalar output & Global pooling & Grid query & Mesh query \\
        \midrule
        Internal energy          & 5.40 & \textbf{5.04} & \underline{5.29} \\
        Top spring force         & 4.53 & \underline{4.21} & \textbf{3.89} \\
        Bottom-left spring force & 5.10 & \underline{4.25} & \textbf{3.90} \\
        Bottom-right spring force& 4.95 & \underline{4.85} & \textbf{4.83} \\
        Contact force            & 3.42 & \textbf{3.20} & \underline{3.37} \\
        Impactor intrusion       & 3.17 & \textbf{2.97} & \underline{2.99} \\
        \midrule
        Mean                     & 4.43 & \underline{4.09} & \textbf{4.04} \\
        \midrule
        Mean final-step field MED (mm) & 0.529 & \textbf{0.463} & \underline{0.469} \\
        \midrule
        \multicolumn{4}{l}{\textit{Computational cost}} \\
        Training time per epoch (s) & \textbf{42.4} & \underline{48.2} & 53.0 \\
        Peak GPU memory (GB)         & \textbf{5.107} & \underline{5.180} & 5.322 \\
        \bottomrule
    \end{tabularx}
\end{table}

\begin{table}[htbp]
    \centering
    \caption{Median absolute relative errors in physical-response metrics on the validation set. Error values are percentages; lower is better. The best value in each row is shown in bold, and the second-best value is underlined.}
    \label{tab:physical_response_error}
    \begin{tabular}{lccc}
        \toprule
        Physical metric & Global pooling & Grid query & Mesh query \\
        \midrule
        Top peak force      & 4.81 & \underline{3.21} & \textbf{2.71} \\
        Bottom-left peak force  & \textbf{2.74} & \underline{2.75} & 3.18 \\
        Bottom-right peak force & \underline{3.25} & \textbf{3.20} & 3.46 \\
        Contact peak force  & \underline{2.43} & \textbf{2.41} & 2.73 \\
        Top impulse         & 2.91 & \textbf{2.27} & \underline{2.40} \\
        Bottom-left impulse  & 3.30 & \textbf{2.37} & \underline{2.43} \\
        Bottom-right impulse & \underline{3.22} & \textbf{2.82} & 3.72 \\
        Contact impulse     & \textbf{2.01} & \underline{2.05} & 2.07 \\
        Final internal energy & \textbf{4.25} & \underline{4.37} & 4.91 \\
        Maximum intrusion   & 2.44 & \textbf{2.10} & \underline{2.35} \\
        \bottomrule
    \end{tabular}
\end{table}

\begin{figure}[htbp]
    \centering
    \includegraphics[width=\textwidth]{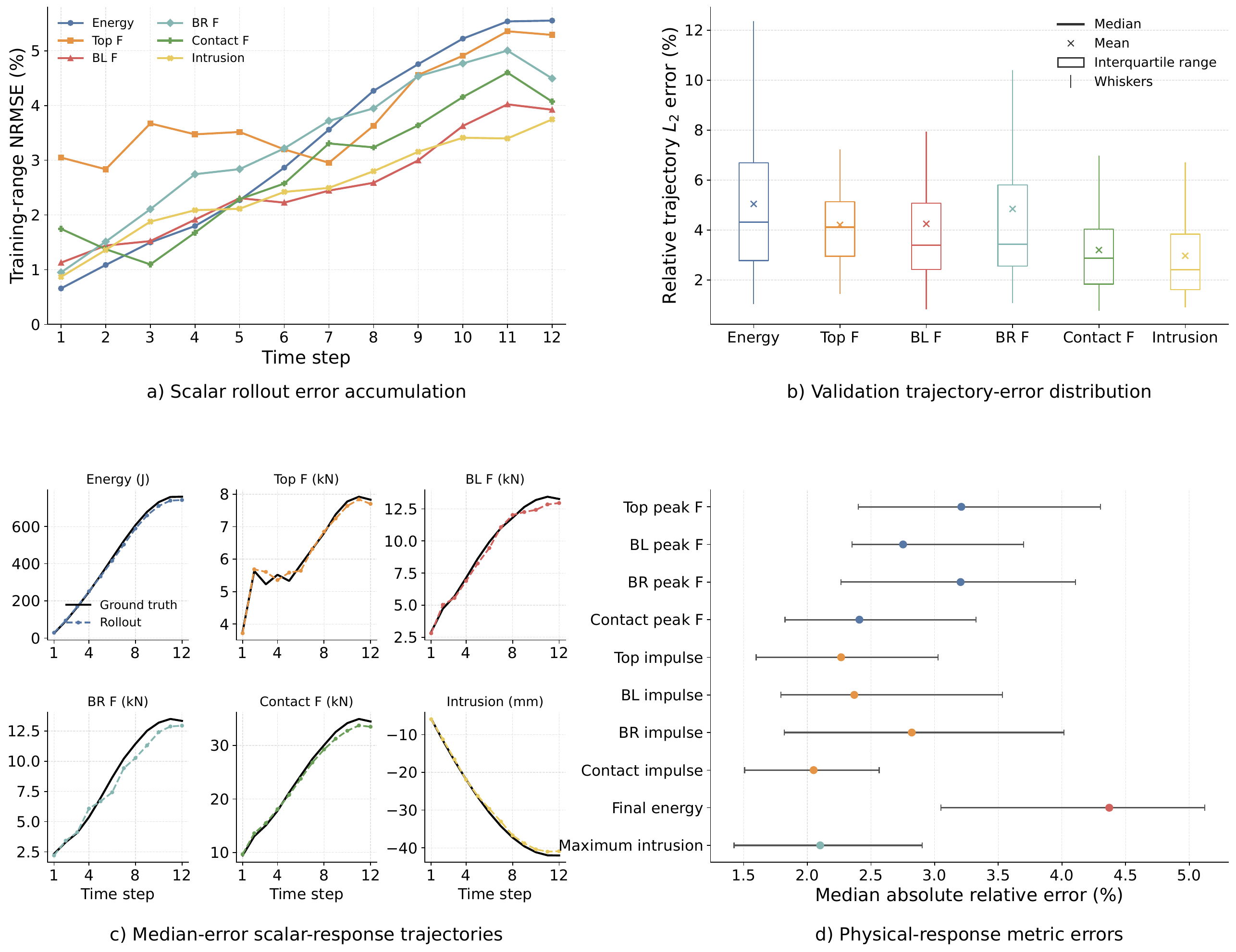}
    \caption{Scalar-response prediction performance. (a) Time-resolved training-range NRMSE over the autoregressive rollout for the six scalar responses. (b) Validation distributions of the relative trajectory $L_2$ error. (c) Ground-truth and predicted trajectories for the median-error validation sample. (d) Median absolute relative errors of the derived physical-response metrics, with error bars showing their variation across the validation samples.}
    \label{fig:scalar_performance}
\end{figure}

Figure~\ref{fig:scalar_performance} quantifies the scalar-prediction accuracy in four ways. First, the error accumulated during the autoregressive rollout is evaluated using the training-range normalised root-mean-square error (NRMSE). For scalar response $k$ at time step $t$, it is defined as
\begin{equation}
    \operatorname{NRMSE}_{k}(t)
    =
    \frac{100}{s^{\mathrm{train}}_{k,\max}-s^{\mathrm{train}}_{k,\min}}
    \sqrt{
        \frac{1}{N_{\mathrm{val}}}
        \sum_{n=1}^{N_{\mathrm{val}}}
        \left(
            \widehat{s}_{n,k,t}-s_{n,k,t}
        \right)^2
    },
\end{equation}
where $N_{\mathrm{val}}$ is the number of validation samples, and $s^{\mathrm{train}}_{k,\max}$ and $s^{\mathrm{train}}_{k,\min}$ are the maximum and minimum values of response $k$ in the training set. NRMSE is a dimensionless error measure used in prior model-evaluation studies~\cite{liemohn2021rmse,bai2019citations}. Normalisation by the fixed training range allows responses with different units and magnitudes to be compared on a common percentage scale without using information from the validation set. Figure~\ref{fig:scalar_performance}(a) shows that the error increases steadily over the rollout, as expected for autoregressive prediction, but remains bounded and does not exhibit explosive accumulation.

Second, Figure~\ref{fig:scalar_performance}(b) presents box plots of the relative trajectory $L_2$ errors across the validation set. The distributions remain within a reasonable range, while contact force and impactor intrusion show lower central errors and narrower spreads than the other responses. Third, Figure~\ref{fig:scalar_performance}(c) compares the ground-truth and predicted trajectories for the validation sample with the median scalar error. The predicted values closely follow the ground truth at every time step for all six scalar responses. A slight underprediction is observed for the bottom-left spring force, but the deviation remains within a reasonable range. Finally, Figure~\ref{fig:scalar_performance}(d) shows the error distributions of the derived crashworthiness indicators. Most validation samples have errors below 5\%, and most indicators have median errors below 3\%. These results demonstrate that the predicted scalar trajectories provide accurate estimates of the engineering quantities used for crashworthiness assessment.

To assess whether the scalar decoder can be transferred across different surrogate models, we train it with four models: ReGUNet, GINO, Transolver, and BAT-NO. ReGUNet, GINO, and BAT-NO use the grid-query decoder because these hierarchical models provide a compact latent representation on which cross-attention can be performed. For ReGUNet, the scalar queries attend to the downsampled graph features, while for GINO and BAT-NO they attend to the regular latent-grid features. Transolver is not hierarchical and does not construct a separate downsampled grid representation, so the mesh-query decoder is applied directly to its mesh-level features. Table~\ref{tab:scalar_model_pair_comparison} shows that ReGUNet and GINO have slightly higher scalar errors but still achieve good prediction accuracy, with mean trajectory errors below 8\%. Although including the predicted scalar responses leads to higher field errors, the overall field-prediction accuracy remains comparable to the field-only variant in most models. Transolver achieves the lowest scalar error for every response, which may indicate that its mesh-level representation is particularly well suited to the mesh-query decoder. Its field accuracy also improves when scalar feedback is included, although its field error remains higher than that of BAT-NO.

The effect of the scalar decoder on field prediction depends on the balance between the additional physical information and the error introduced by the predicted scalar responses. Feeding the scalar features back into the model provides complementary information about the global and local structural response, which can improve the subsequent field prediction. However, scalar-prediction errors are also propagated through the autoregressive rollout and may outweigh this benefit when they become sufficiently large. Transolver demonstrates the beneficial case, where its low scalar errors lead to improved field accuracy. For GINO and BAT-NO, the differences between the models with and without scalar feedback are small and are considered acceptable given the additional scalar outputs. These results show that the proposed decoder is architecture-flexible and can be attached to different surrogate models by querying an appropriate latent or mesh-level representation.

\begin{table}[htbp]
    \centering
    \caption{Comparison of scalar-response and displacement-field rollout accuracy on the validation set. Scalar-response values are mean relative trajectory errors (\%); field values are mean final-step MED values (mm). Lower is better. The best value in each row is shown in bold, and the second-best value is underlined. ReGUNet, GINO, and BAT-NO use the grid-query decoder, while Transolver uses the mesh-query decoder.}
    \label{tab:scalar_model_pair_comparison}
    \begin{tabular}{lcccc}
        \toprule
        Metric & ReGUNet & GINO & Transolver & BAT-NO \\
        \midrule
        Internal energy              & 7.81  & 8.29  & \textbf{2.40} & \underline{5.04} \\
        Top spring force             & 8.54  & 4.84  & \textbf{3.46} & \underline{4.21} \\
        Bottom-left spring force     & 10.80 & 5.93  & \textbf{2.95} & \underline{4.25} \\
        Bottom-right spring force    & 9.57  & 6.80  & \textbf{3.57} & \underline{4.85} \\
        Contact force                & 5.46  & 5.09  & \textbf{2.02} & \underline{3.20} \\
        Impactor intrusion           & 4.40  & 4.73  & \textbf{1.93} & \underline{2.97} \\
        \midrule
        Mean scalar trajectory error & 7.76  & 5.95  & \textbf{2.72} & \underline{4.09} \\
        \midrule
        Field MED with scalar decoder    & 1.438 & 0.675 & \underline{0.630} & \textbf{0.463} \\
        Field MED without scalar decoder & 1.106 & \underline{0.669} & 0.742 & \textbf{0.451} \\
        \bottomrule
    \end{tabular}
\end{table}

The preceding experiments represent two reference configurations. The baseline and ablation studies use the untuned field-only BAT-NO, which achieves a validation final-step MED of $0.451$~mm. The scalar-decoder study uses the untuned jointly trained BAT-NO, for which the selected grid-query decoder achieves a field MED of $0.463$~mm and a mean scalar-trajectory error of 4.09\%. The following section evaluates the final BAT-NO configuration after hyperparameter tuning, with joint field and scalar prediction enabled.

\subsection{Final Model Configuration and Hyperparameter Selection}
\label{subsec:final_model_configuration}

Hyperparameter selection is performed on Case SLB using its training and validation sets. Candidate configurations are ranked by their mean validation final-step MED. The main text presents the results for three important architecture hyperparameters: the number of boundary-condition-attention slices, the number of FNO layers, and the latent-grid resolution. The remaining architecture and training-parameter screenings are reported in Appendix~\ref{app:hyperparameter_tuning}. All training experiments are conducted on an NVIDIA Quadro RTX 5000 GPU with 16~GB of memory.

\begin{figure}[htbp]
\centering
\begin{minipage}[t]{0.48\textwidth}
    \centering
    \includegraphics[width=\linewidth]{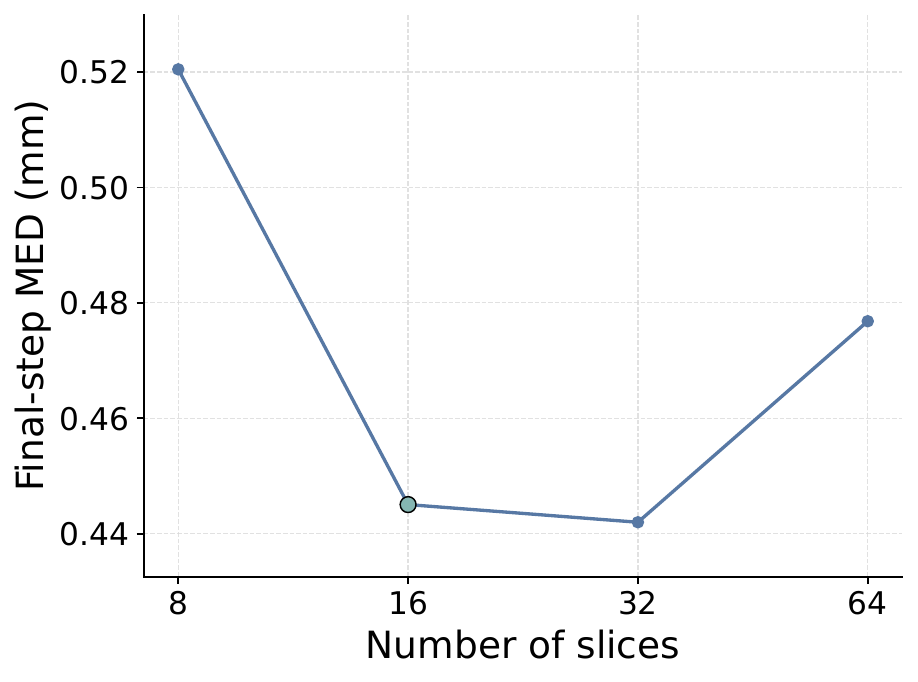}
    \small (a)
\end{minipage}
\hfill
\begin{minipage}[t]{0.48\textwidth}
    \centering
    \includegraphics[width=\linewidth]{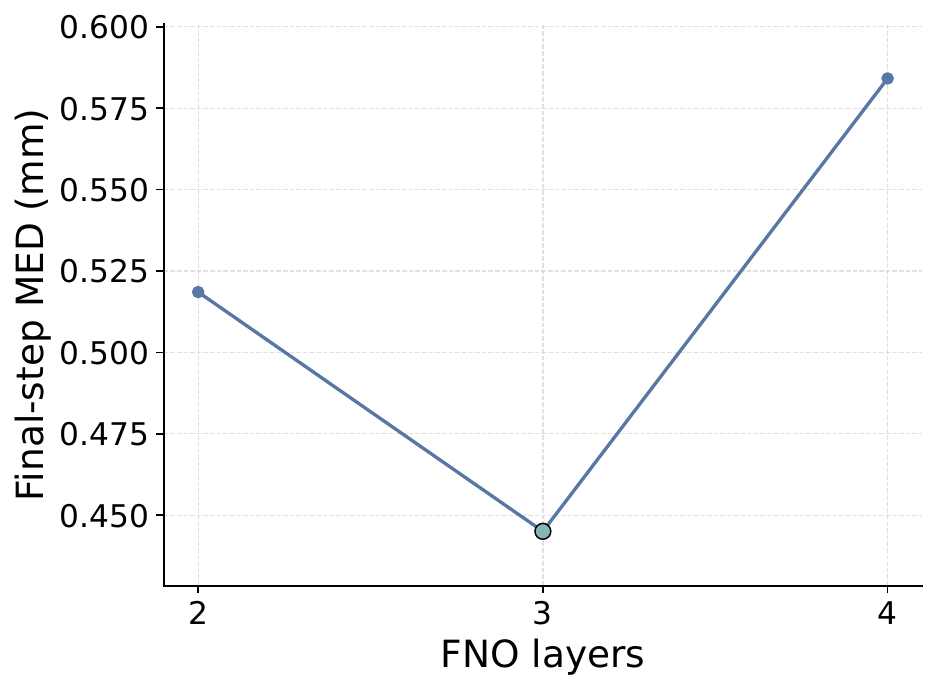}
    \small (b)
\end{minipage}
\caption{Validation final-step MED for (a) the number of physics-attention slices and (b) the FNO depth.}
\label{fig:key-hyperparameter-screening}
\end{figure}

As shown in Fig.~\ref{fig:key-hyperparameter-screening}(a), for the number of boundary-condition-attention slices, increasing the number of slices from 8 to 16 substantially reduces the error. While 32 slices provide only a marginal further reduction from 0.445 to 0.442~mm, they require greater GPU-memory consumption and longer training time. Sixteen slices are therefore retained as the cost-aware configuration, as increasing the count to 64 degrades accuracy. The number of FNO layers is selected to be 3 because it produces the lowest validation error within the FNO-depth search. 

\begin{figure}[htbp]
\centering
\begin{minipage}[t]{0.48\textwidth}
    \centering
    \includegraphics[width=\linewidth]{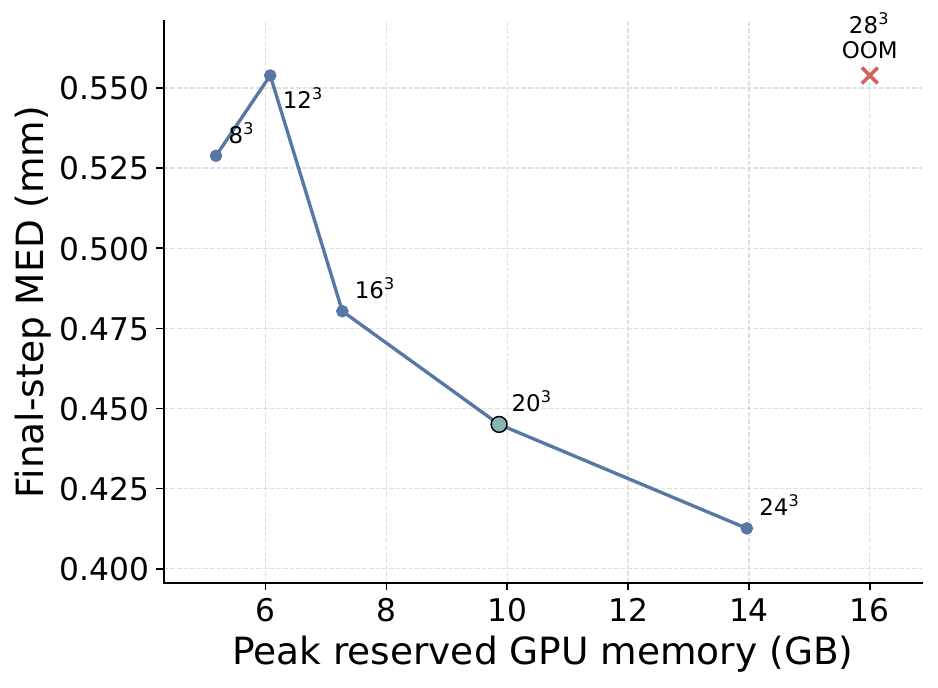}
    \small (a)
\end{minipage}
\hfill
\begin{minipage}[t]{0.48\textwidth}
    \centering
    \includegraphics[width=\linewidth]{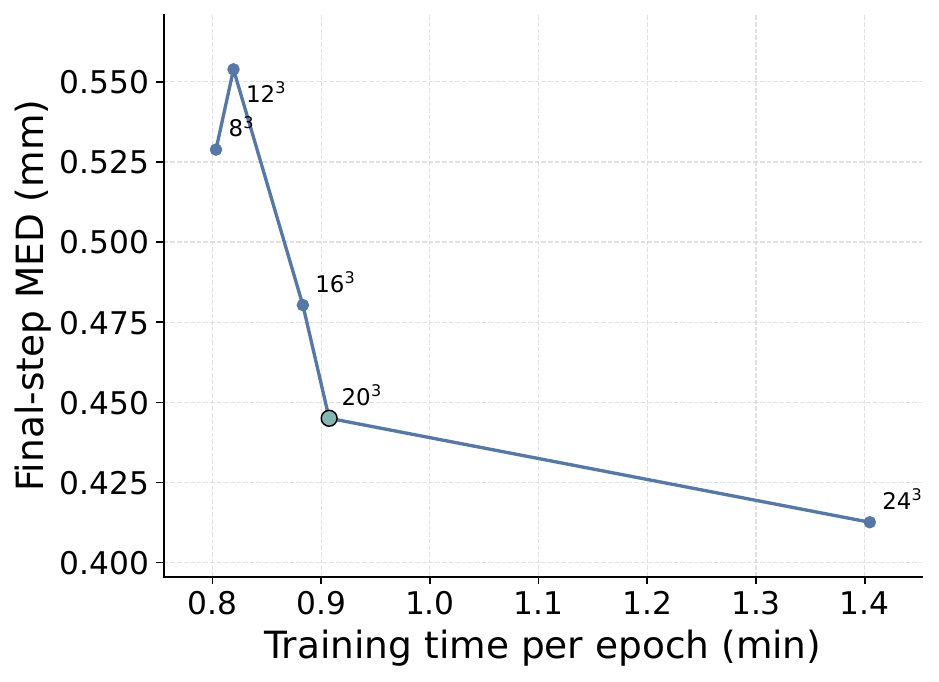}
    \small (b)
\end{minipage}
\caption{Accuracy--efficiency trade-off for different latent-grid resolutions: (a) validation error versus peak GPU memory and (b) validation error versus training time.}
\label{fig:grid-resolution-efficiency}
\end{figure}

The latent-grid resolution determines the spatial capacity of the regular-grid representation, but its computational and memory requirements increase rapidly with resolution. The resulting accuracy--efficiency trade-off is visualised in Fig.~\ref{fig:grid-resolution-efficiency}. Although the $24^3$ grid achieves the lowest validation error among the feasible configurations and the $28^3$ grid exceeds the available GPU memory, the $20^3$ grid provides a more balanced trade-off between prediction accuracy and computational efficiency. It is therefore selected for the final model. For applications involving more complex input fields, a finer latent grid may provide additional representational capacity when stronger GPU resources are available. 

\ref{app:hyperparameter_tuning} provides the complete tuning procedure, search ranges, and additional results. These include further architecture parameters, such as the number of attention heads, pre-grid message-passing steps, and slice-transformer layers, together with training parameters including the learning rate, batch size, scalar-loss weight, attention dropout, weight decay, learning-rate schedule, and maximum predicted-feedback probability used for scheduled sampling. The resulting final architecture and training settings are summarised in Table~\ref{tab:bat-no-final-hyperparameters}.

\begin{table}[htbp]
\centering
\caption{Final hyperparameter settings used for BAT-NO.}
\label{tab:bat-no-final-hyperparameters}
\small
\begin{tabular}{lclc}
\toprule
\multicolumn{2}{c}{\textbf{Model architecture}} &
\multicolumn{2}{c}{\textbf{Training}} \\
\cmidrule(lr){1-2}\cmidrule(lr){3-4}
Hyperparameter & Setting & Hyperparameter & Setting \\
\midrule
Latent grid resolution & $20^3$ & Learning rate & $3\times10^{-4}$ \\
Fourier modes & 4 & Batch size & 2 \\
FNO layers & 3 & Scalar-loss weight & 0.10 \\
Number of slices & 16 & Learning-rate schedule & Constant \\
Attention heads & 4 & Attention dropout & 0 \\
Pre-grid MP steps & 4 & Weight decay & 0 \\
Slice-transformer layers & 1 & Maximum predicted feedback & 0.75 \\
\bottomrule
\end{tabular}
\end{table}

\begin{figure}[htbp]
    \centering
    \includegraphics[width=0.92\textwidth]{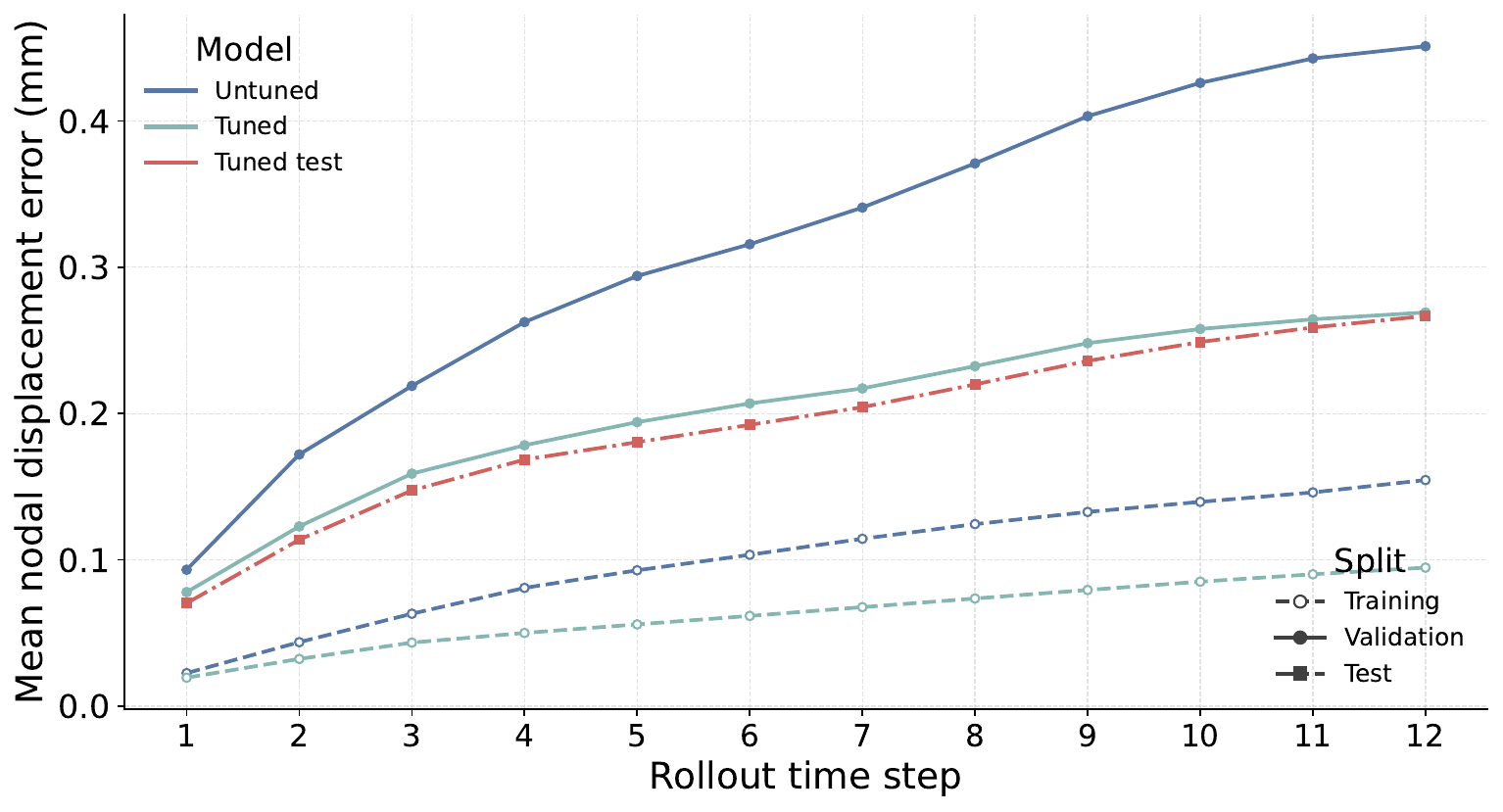}
    \caption{Mean nodal Euclidean displacement error over the autoregressive rollout for the untuned field-only BAT-NO and the final tuned field-and-scalar BAT-NO. Training and validation curves are shown for both configurations, together with the tuned-model test error on $300$ test samples.}
    \label{fig:batno_tuning_rollout}
\end{figure}

As shown in Fig.~\ref{fig:batno_tuning_rollout}, the final tuned BAT-NO model reduces the rollout error across the full prediction trajectory compared with the untuned field-only configuration used in the baseline and ablation studies. The improvement becomes more pronounced at later time steps, indicating that the final model configuration reduces error accumulation during autoregressive prediction. The untuned reference model achieves a validation final-step MED of $0.451$~mm, while the final tuned model achieves $0.269$~mm. The tuned model also achieves a comparable error of $0.267$~mm on the test set, which contains $300$ unseen samples within the investigated design space. The validation and test results of the final tuned model were obtained using fully autoregressive evaluation, in which the predicted incremental-displacement field and the six predicted scalar responses were fed back as inputs at subsequent time steps. This suggests that the tuned configuration improves generalisation rather than simply fitting the validation set. The training curves remain lower than the validation and test curves, as expected, but the gap does not increase sharply over time for the tuned model, indicating more stable rollout behaviour. A full-trajectory field and error visualisation for the median-error test sample is provided in Fig.~\ref{fig:test_median_rollout_z_field} in \ref{app:full_trajectory_visualisation}.

Table~\ref{tab:tuned_scalar_performance} reports the scalar-response accuracy of the final tuned BAT-NO under fully autoregressive evaluation. At each time step, both the predicted incremental-displacement field and the six predicted scalar responses are fed back into the model. The tuned model achieves a mean relative scalar-trajectory error of $2.48\%$ on the validation set and $2.46\%$ on the independent test set.

\begin{table}[htbp]
    \centering
    \caption{Field and scalar-response accuracy of the final tuned BAT-NO under fully autoregressive evaluation. Scalar-response values are mean relative trajectory $L_2$ errors (\%).}
    \label{tab:tuned_scalar_performance}
    \begin{tabular}{lcc}
        \toprule
        Metric & Validation & Test \\
        \midrule
        Internal energy              & 2.14 & 2.19 \\
        Top spring force             & 3.04 & 3.28 \\
        Bottom-left spring force     & 3.64 & 3.17 \\
        Bottom-right spring force    & 2.82 & 2.85 \\
        Contact force                & 1.77 & 1.84 \\
        Impactor intrusion           & 1.48 & 1.46 \\
        \midrule
        Mean scalar trajectory error & 2.48 & 2.46 \\
        Final-step field MED (mm)    & 0.269 & 0.267 \\
        \bottomrule
    \end{tabular}
\end{table}

Compared with the untuned BAT-NO result of $4.09\%$ reported in Table~\ref{tab:scalar_model_pair_comparison}, the final selected configuration reduces the validation mean scalar-trajectory error by $39.4\%$. The similar validation and test errors indicate that the improvement is retained across the independently generated test simulations.

\section{Conclusion}
\label{sec:conclusion}

This work introduced BAT-NO for component-level crashworthiness prediction under variations in geometry, loading, and support conditions. In the simulation framework, geometry was varied through mesh morphing, loading was varied by translating the impactor vertically and changing its initial velocity, and support conditions were varied by changing the stiffnesses of elastic supports at the support regions. BAT-NO combines mesh message passing with latent-grid operator learning and uses two complementary condition pathways. Slice attention captures global interactions among physically related regions, while the direct boundary-to-grid projection preserves the local spatial distribution of support and loading information. Temporal recurrence is used to retain historical information through the autoregressive crash trajectory, limiting error accumulation.

Across the three case studies, BAT-NO achieved the lowest mean final-step MED among the evaluated graph, neural-operator, and transformer baselines. The mean validation errors were $0.1318$~mm for shape variation, $0.1742$~mm for combined shape and loading variation, and $0.4510$~mm when shape, loading, and support conditions varied simultaneously. In the most challenging case, BAT-NO reduced the mean error by 32.6\% relative to the second-best model. The ablation study showed that slice-based conditioning, pre-grid message passing, temporal recurrence, and direct boundary-condition projection each contributed to validation accuracy. The results indicate that the slice and direct-projection pathways provide complementary global and local descriptions of the boundary conditions, reducing the prediction error of validation samples. Following hyperparameter tuning for Case SLB, the final jointly trained field-and-scalar model achieved a validation error of $0.269$~mm and a comparable error of $0.267$~mm on the independently generated test set of 300 simulations.

The attention-based scalar decoder enabled joint prediction of displacement fields and six crashworthiness-response trajectories. The selected grid-query decoder achieved a mean relative scalar-trajectory error of 2.46\%. All ten evaluated physical-response indicators had median errors below 5\%, and seven had median errors below 3\%. Applying the decoder to ReGUNet, GINO, Transolver, and BAT-NO further showed that it can query either compact latent-grid features or mesh-level features, making it reusable across different surrogate architectures. The results also show that feeding predicted scalar responses back into the rollout does not necessarily improve field accuracy, since additional physical response information can introduce scalar-prediction errors into later time steps.

This study is limited to simulation data from one component-level side-impact configuration, with neighbouring-structure effects approximated through independently varied elastic supports. Future work should evaluate the approach on other vehicle components and larger assemblies, incorporate broader variations in material, joint, and contact behaviour, and assess generalisation to conditions outside the training design space. Extension to higher-resolution meshes and more complex crash scenarios will also require further improvement in the efficiency of the latent-grid and attention operations. Overall, the results show that combining local and global boundary-condition representations is effective for learning transient crash responses across expanded component-level design spaces.

\section*{Acknowledgements}
The authors acknowledge funding support from UKRI (UKRI221: AI-Driven Design for Forming High-Performance Vehicle Parts), as well as PhD scholarships from Imperial College London. For the purpose of open access, the authors have applied a Creative Commons Attribution (CC BY) license to any Author Accepted Manuscript version arising.

\section*{Declaration of Competing Interest}
The authors declare that they have no known competing financial interests or personal relationships that could have appeared to influence the work reported in this paper.

\section*{Declaration of generative AI and AI-assisted technologies in the manuscript preparation process.}
During the preparation of this work the authors used ChatGPT in order to refine language. After using this tool, the authors reviewed and edited the content as needed and take full responsibility for the content of the published article.

\bibliographystyle{elsarticle-num}
\bibliography{references}

@inproceedings{li2021fno,
  title     = {Fourier Neural Operator for Parametric Partial Differential Equations},
  author    = {Li, Zongyi and Kovachki, Nikola Borislavov and Azizzadenesheli, Kamyar and Liu, Burigede and Bhattacharya, Kaushik and Stuart, Andrew M. and Anandkumar, Anima},
  booktitle = {International Conference on Learning Representations},
  year      = {2021},
  url       = {https://openreview.net/forum?id=c8P9NQVtmnO}
}

@article{li2020gno,
  title   = {Neural Operator: Graph Kernel Network for Partial Differential Equations},
  author  = {Li, Zongyi and Kovachki, Nikola and Azizzadenesheli, Kamyar and Liu, Burigede and Bhattacharya, Kaushik and Stuart, Andrew and Anandkumar, Anima},
  journal = {arXiv preprint arXiv:2003.03485},
  year    = {2020},
  url     = {https://arxiv.org/abs/2003.03485}
}

@article{wen2022ufno,
  title   = {{U-FNO}: An Enhanced Fourier Neural Operator-Based Deep-Learning Model for Multiphase Flow},
  author  = {Wen, Gege and Li, Zongyi and Azizzadenesheli, Kamyar and Anandkumar, Anima and Benson, Sally M.},
  journal = {Advances in Water Resources},
  volume  = {163},
  pages   = {104180},
  year    = {2022}
}

@article{li2022geofno,
  title   = {Fourier Neural Operator with Learned Deformations for {PDEs} on General Geometries},
  author  = {Li, Zongyi and Huang, Daniel Zhengyu and Liu, Burigede and Anandkumar, Anima},
  journal = {Journal of Machine Learning Research},
  volume  = {24},
  number  = {388},
  pages   = {1--26},
  year    = {2023}
}

@inproceedings{li2023gino,
  title     = {Geometry-Informed Neural Operator for Large-Scale 3D PDEs},
  author    = {Li, Zongyi and Kovachki, Nikola B. and Choy, Christopher and Li, Boyi and Kossaifi, Jean and Otta, Shourya Prakash and Nabian, Mohammad Amin and Stadler, Maximilian and Hundt, Christian and Azizzadenesheli, Kamyar and Anandkumar, Anima},
  booktitle = {Advances in Neural Information Processing Systems},
  volume    = {36},
  year      = {2023},
  url       = {https://proceedings.neurips.cc/paper_files/paper/2023/hash/70518ea42831f02afc3a2828993935ad-Abstract-Conference.html}
}

@article{liu2025ginot,
  title   = {Geometry-Informed Neural Operator Transformer},
  author  = {Liu, Qibang and Zhong, Vincient and Meidani, Hadi and Abueidda, Diab W. and Koric, Seid and Geubelle, Philippe H.},
  journal = {arXiv preprint arXiv:2504.19452},
  year    = {2025}
}

@article{kim2026eqgino,
  title   = {{EqGINO}: Equivariant Geometry-Informed Fourier Neural Operators for 3D PDEs},
  author  = {Kim, Sungwon and Song, Juho and Shin, Seungmin and Cho, Guimok and Kim, Sangkook and Park, Chanyoung},
  journal = {arXiv preprint arXiv:2606.03260},
  year    = {2026}
}

@inproceedings{wu2024transolver,
  title     = {Transolver: A Fast Transformer Solver for {PDEs} on General Geometries},
  author    = {Wu, Haixu and Luo, Huakun and Wang, Haowen and Wang, Jianmin and Long, Mingsheng},
  booktitle = {Proceedings of the 41st International Conference on Machine Learning},
  series    = {Proceedings of Machine Learning Research},
  volume    = {235},
  pages     = {53681--53705},
  year      = {2024},
  publisher = {PMLR},
  url       = {https://proceedings.mlr.press/v235/wu24r.html}
}

@inproceedings{luo2025transolverpp,
  title     = {Transolver++: An Accurate Neural Solver for {PDEs} on Million-Scale Geometries},
  author    = {Luo, Huakun and Wu, Haixu and Zhou, Hang and Xing, Lanxiang and Di, Yichen and Wang, Jianmin and Long, Mingsheng},
  booktitle = {International Conference on Machine Learning},
  year      = {2025}
}

@article{wu2026transolver3,
  title   = {Transolver-3: Scaling Up Transformer Solvers to Industrial-Scale Geometries},
  author  = {Wu, Haixu and others},
  journal = {arXiv preprint arXiv:2602.04940},
  year    = {2026}
}

@inproceedings{kashi2024lpfno,
  title     = {Learning the Boundary-to-Domain Mapping Using Lifting Product Fourier Neural Operators for Partial Differential Equations},
  author    = {Kashi, Aditya and Daw, Arka and Meena, Muralikrishnan Gopalakrishnan and Lu, Hao},
  booktitle = {{ICML} 2024 Workshops: {AI4Science}},
  year      = {2024},
  url       = {https://mlanthology.org/icmlw/2024/kashi2024icmlw-learning/}
}

@article{boya2025pinto,
  title   = {{PINTO}: Physics-Informed Transformer Neural Operator for Learning Generalized Solutions of Partial Differential Equations for Any Initial and Boundary Condition},
  author  = {Boya, Sumanth Kumar and Subramani, Deepak N.},
  journal = {Computer Physics Communications},
  volume  = {315},
  pages   = {109702},
  year    = {2025},
  doi     = {10.1016/j.cpc.2025.109702}
}

@article{mousavi2026boundary,
  title   = {Imposing Boundary Conditions on Neural Operators via Learned Function Extensions},
  author  = {Mousavi, Sepehr and Mishra, Siddhartha and De Lorenzis, Laura},
  journal = {arXiv preprint arXiv:2602.04923},
  year    = {2026},
  doi     = {10.48550/arXiv.2602.04923}
}

@article{adams2025geotransolver,
  title   = {{GeoTransolver}: Learning Physics on Irregular Domains Using Multi-Scale Geometry Aware Physics Attention Transformer},
  author  = {Adams, Corey and Ranade, Rishikesh and Cherukuri, Ram and Choudhry, Sanjay},
  journal = {arXiv preprint arXiv:2512.20399},
  year    = {2025},
  doi     = {10.48550/arXiv.2512.20399}
}

@inproceedings{mousavi2025rigno,
  title     = {{RIGNO}: A Graph-Based Framework for Robust and Accurate Operator Learning for {PDEs} on Arbitrary Domains},
  author    = {Mousavi, Sepehr and Wen, Shizheng and Lingsch, Levi and Herde, Maximilian and Raoni{\'c}, Bogdan and Mishra, Siddhartha},
  booktitle = {Advances in Neural Information Processing Systems},
  volume    = {38},
  year      = {2025},
  url       = {https://github.com/camlab-ethz/rigno}
}

@article{wen2025gaot,
  title   = {Geometry Aware Operator Transformer as an Efficient and Accurate Neural Surrogate for {PDEs} on Arbitrary Domains},
  author  = {Wen, Shizheng and Kumbhat, Arsh and Lingsch, Levi and Mousavi, Sepehr and Zhao, Yizhou and Chandrashekar, Praveen and Mishra, Siddhartha},
  journal = {arXiv preprint arXiv:2505.18781},
  year    = {2025},
  doi     = {10.48550/arXiv.2505.18781}
}

@article{li2026maskmorph,
  title   = {Mask-Morph Graph U-Net: A Generalisable Mesh-Based Surrogate for Crashworthiness Field Prediction under Large Geometric Variation},
  author  = {Li, Haoran and Lehrer, Tobias and Zhao, Yingxue and Zhou, Haosu and Stocker, Philipp and Pfaff, Tobias and Wagner, Marcus and Li, Nan},
  journal = {arXiv preprint arXiv:2605.15231},
  year    = {2026}
}

@article{akhare2026crash,
  title   = {High-Fidelity Industrial Crash Dynamics Prediction via Geometry-Aware Operator Learning with Memory-Efficient Low-Rank Attention},
  author  = {Akhare, Deepak and Nabian, Mohammad Amin and Adams, Corey and Chavare, Sudeep and Choudhry, Sanjay},
  journal = {arXiv preprint arXiv:2605.27758},
  year    = {2026},
  doi     = {10.48550/arXiv.2605.27758}
}

@article{elrefaie2026carcrashnet,
  title   = {{CarCrashNet}: A Large-Scale Dataset and Hierarchical Neural Solver for Data-Driven Structural Crash Simulation},
  author  = {Elrefaie, Mohamed and Shu, Dule and Klenk, Matthew and Ahmed, Faez},
  journal = {arXiv preprint arXiv:2605.07098},
  year    = {2026},
  doi     = {10.48550/arXiv.2605.07098}
}

@inproceedings{pfaff2021meshgraphnets,
  title     = {Learning Mesh-Based Simulation with Graph Networks},
  author    = {Pfaff, Tobias and Fortunato, Meire and Sanchez-Gonzalez, Alvaro and Battaglia, Peter W.},
  booktitle = {International Conference on Learning Representations},
  year      = {2021},
  url       = {https://openreview.net/forum?id=roNqYL0_XP}
}

@article{fortunato2022multiscale,
  title   = {MultiScale MeshGraphNets},
  author  = {Fortunato, Meire and Pfaff, Tobias and Wirnsberger, Peter and Pritzel, Alexander and Battaglia, Peter W.},
  journal = {arXiv preprint arXiv:2210.00612},
  year    = {2022},
  url     = {https://arxiv.org/abs/2210.00612}
}

@article{nabian2024xmeshgraphnet,
  title   = {{X-MeshGraphNet}: Scalable Multi-Scale Graph Neural Networks for Physics Simulation},
  author  = {Nabian, Mohammad Amin and Liu, Chang and Ranade, Rishikesh and Choudhry, Sanjay},
  journal = {arXiv preprint arXiv:2411.17164},
  year    = {2024},
  doi     = {10.48550/arXiv.2411.17164}
}

@article{li2026regunet,
  title   = {A Graph Neural Network Surrogate Model for Mesh-Based Crashworthiness Prediction of Vehicle Panel Components},
  author  = {Li, Haoran and Zhao, Yingxue and Zhou, Haosu and Pfaff, Tobias and Li, Nan},
  journal = {Results in Engineering},
  volume  = {30},
  pages   = {110925},
  year    = {2026},
  doi     = {10.1016/j.rineng.2026.110925}
}

@article{lu2021deeponet,
  title   = {Learning Nonlinear Operators via {DeepONet} Based on the Universal Approximation Theorem of Operators},
  author  = {Lu, Lu and Jin, Pengzhan and Pang, Guofei and Zhang, Zhongqiang and Karniadakis, George Em},
  journal = {Nature Machine Intelligence},
  volume  = {3},
  number  = {3},
  pages   = {218--229},
  year    = {2021},
  doi     = {10.1038/s42256-021-00302-5}
}

@inproceedings{alkin2024upt,
  title     = {Universal Physics Transformers: A Framework For Efficiently Scaling Neural Operators},
  author    = {Alkin, Benedikt and F{\"u}rst, Andreas and Schmid, Simon and Gruber, Lukas and Holzleitner, Markus and Brandstetter, Johannes},
  booktitle = {Advances in Neural Information Processing Systems},
  volume    = {37},
  year      = {2024},
  doi       = {10.52202/079017-0793}
}

@article{nabian2026automotive,
  title   = {Automotive Crash Dynamics Modeling Accelerated with Machine Learning},
  author  = {Nabian, Mohammad Amin and Chavare, Sudeep and Akhare, Deepak and Ranade, Rishikesh and Cherukuri, Ram and Tadepalli, Srinivas},
  journal = {SAE Technical Paper},
  number  = {2026-01-0568},
  year    = {2026},
  doi     = {10.4271/2026-01-0568}
}

@article{curtosi2026crash,
  title   = {Crash Assessment via Mesh-Based Graph Neural Networks and Physics-Aware Attention},
  author  = {Curtosi, Gabriel and Ruiz Ruiz, Carlos Manuel and Cavaliere, Fabiola and Larr{\'a}yoz Izcara, Xabier},
  journal = {arXiv preprint arXiv:2605.11784},
  year    = {2026},
  doi     = {10.48550/arXiv.2605.11784}
}

@article{iparraguirre2026mgnt,
  title   = {{MeshGraphNet-Transformer}: Scalable Mesh-Based Learned Simulation for Solid Mechanics},
  author  = {Iparraguirre, Mikel M. and Alfaro, Iciar and Gonzalez, David and Cueto, Elias},
  journal = {arXiv preprint arXiv:2601.23177},
  year    = {2026}
}

@article{mckay1979latin,
  title   = {A Comparison of Three Methods for Selecting Values of Input Variables in the Analysis of Output from a Computer Code},
  author  = {McKay, Michael D. and Beckman, Richard J. and Conover, William J.},
  journal = {Technometrics},
  volume  = {21},
  number  = {2},
  pages   = {239--245},
  year    = {1979},
  doi     = {10.1080/00401706.1979.10489755}
}

@inproceedings{jaegle2021perceiver,
  title     = {Perceiver: General Perception with Iterative Attention},
  author    = {Jaegle, Andrew and Gimeno, Felix and Brock, Andrew and Zisserman, Andrew and Vinyals, Oriol and Carreira, Joao},
  booktitle = {Proceedings of the 38th International Conference on Machine Learning},
  year      = {2021}
}

@article{jaegle2021perceiverio,
  title   = {Perceiver IO: A General Architecture for Structured Inputs and Outputs},
  author  = {Jaegle, Andrew and Borgeaud, Sebastian and Alayrac, Jean-Baptiste and others},
  journal = {arXiv preprint arXiv:2107.14795},
  year    = {2021}
}

@article{liemohn2021rmse,
  title   = {{RMSE} Is Not Enough: Guidelines to Robust Data--Model Comparisons for Magnetospheric Physics},
  author  = {Liemohn, Michael W. and Shane, Alexander D. and Azari, Abigail R. and Petersen, Alexander K. and Swiger, Benjamin M. and Mukhopadhyay, Ayan},
  journal = {Journal of Atmospheric and Solar-Terrestrial Physics},
  volume  = {218},
  pages   = {105624},
  year    = {2021},
  doi     = {10.1016/j.jastp.2021.105624}
}

@article{bai2019citations,
  title   = {Predicting the Citations of Scholarly Paper},
  author  = {Bai, Xiaomei and Zhang, Fuli and Lee, Ivan},
  journal = {Journal of Informetrics},
  volume  = {13},
  number  = {1},
  pages   = {407--418},
  year    = {2019},
  doi     = {10.1016/j.joi.2019.01.010}
}

@article{attar2021hfqcnn,
  title   = {Deformation and Thinning Field Prediction for {HFQ}-Formed Panel Components Using Convolutional Neural Networks},
  author  = {Attar, Hamid Reza and Zhou, Haosu and Li, Nan},
  journal = {IOP Conference Series: Materials Science and Engineering},
  volume  = {1157},
  number  = {1},
  pages   = {012079},
  year    = {2021},
  doi     = {10.1088/1757-899X/1157/1/012079}
}

@article{xu2022supermeshing,
  title   = {{SuperMeshing}: A New Deep Learning Architecture for Increasing the Mesh Density of Physical Fields in Metal Forming Numerical Simulation},
  author  = {Xu, Qingfeng and Nie, Zhenguo and Xu, Handing and Zhou, Haosu and Attar, Hamid Reza and Li, Nan and Xie, Fugui and Liu, Xin-Jun},
  journal = {Journal of Applied Mechanics},
  volume  = {89},
  number  = {1},
  pages   = {011002},
  year    = {2022},
  doi     = {10.1115/1.4052195}
}

@article{ding2024cnnconstitutive,
  title   = {Integrating Convolutional Neural Network and Constitutive Model for Rapid Prediction of Stress-Strain Curves in Fibre Reinforced Polymers: A Generalisable Approach},
  author  = {Ding, Zerong and Attar, Hamid Reza and Wang, Hongyan and Liu, Haibao and Li, Nan},
  journal = {Materials \& Design},
  volume  = {241},
  pages   = {112849},
  year    = {2024},
  doi     = {10.1016/j.matdes.2024.112849}
}

@article{attar2023implicit,
  title   = {Implicit Neural Representations of Sheet Stamping Geometries with Small-Scale Features},
  author  = {Attar, Hamid Reza and Foster, Alistair and Li, Nan},
  journal = {Engineering Applications of Artificial Intelligence},
  volume  = {123},
  pages   = {106482},
  year    = {2023},
  doi     = {10.1016/j.engappai.2023.106482}
}

@article{attar2023stamping,
  title   = {Development of a Deep Learning Platform for Sheet Stamping Geometry Optimisation under Manufacturing Constraints},
  author  = {Attar, Hamid Reza and Foster, Alistair and Li, Nan},
  journal = {Engineering Applications of Artificial Intelligence},
  volume  = {123},
  pages   = {106295},
  year    = {2023},
  doi     = {10.1016/j.engappai.2023.106295}
}

@article{li2024cnncrash,
  title   = {An Integrated Convolutional Neural Network-Based Surrogate Model for Crashworthiness Performance Prediction of Hot-Stamped Vehicle Panel Components},
  author  = {Li, Haoran and Zhou, Haosu and Li, Nan},
  journal = {MATEC Web of Conferences},
  volume  = {401},
  pages   = {03013},
  year    = {2024},
  doi     = {10.1051/matecconf/202440103013}
}

@article{zhou2025multilevelgcnn,
  title   = {A Multi-Level Graph-Based Surrogate Model for Real-Time High-Fidelity Sheet Forming Simulations},
  author  = {Zhou, Haosu and Zhao, Yingxue and Li, Haoran and Pfaff, Tobias and Li, Nan},
  journal = {Advanced Engineering Informatics},
  volume  = {66},
  pages   = {103458},
  year    = {2025},
  doi     = {10.1016/j.aei.2025.103458}
}

@article{zhao2026rugnn,
  title   = {Recurrent {U-Net}-Based Graph Neural Network ({RUGNN}) for Accurate Deformation Predictions in Sheet Material Forming},
  author  = {Zhao, Yingxue and Chen, Qianyi and Li, Haoran and Zhou, Haosu and Attar, Hamid Reza and Pfaff, Tobias and Wu, Tailin and Li, Nan},
  journal = {Advanced Engineering Informatics},
  volume  = {69},
  pages   = {104021},
  year    = {2026},
  doi     = {10.1016/j.aei.2025.104021}
}

@article{zhao2024gnnreview,
  title   = {A Review of Graph Neural Network Applications in Mechanics-Related Domains},
  author  = {Zhao, Yingxue and Li, Haoran and Zhou, Haosu and Attar, Hamid Reza and Pfaff, Tobias and Li, Nan},
  journal = {Artificial Intelligence Review},
  volume  = {57},
  pages   = {315},
  year    = {2024},
  doi     = {10.1007/s10462-024-10931-y}
}

@article{zhao2025batterybox,
  title   = {Rapid Prediction of Material Deformation in Hot Stamping of Battery Box Geometries Using Graph Neural Network},
  author  = {Zhao, Yingxue and Li, Haoran and Zhou, Haosu and Attar, Hamid Reza and Pfaff, Tobias and Li, Nan},
  journal = {Journal of Physics: Conference Series},
  volume  = {3104},
  number  = {1},
  pages   = {012057},
  year    = {2025},
  doi     = {10.1088/1742-6596/3104/1/012057}
}

@article{attar2021rapid,
  title   = {Rapid Feasibility Assessment of Components to Be Formed through Hot Stamping: A Deep Learning Approach},
  author  = {Attar, Hamid Reza and Zhou, Haosu and Foster, Alistair and Li, Nan},
  journal = {Journal of Manufacturing Processes},
  volume  = {68},
  pages   = {1650--1671},
  year    = {2021},
  doi     = {10.1016/j.jmapro.2021.06.011}
}

@article{zhou2022imagestamping,
  title   = {A Study on Using Image-Based Machine Learning Methods to Develop Surrogate Models of Stamp Forming Simulations},
  author  = {Zhou, Haosu and Xu, Qingfeng and Nie, Zhenguo and Li, Nan},
  journal = {Journal of Manufacturing Science and Engineering},
  volume  = {144},
  number  = {2},
  pages   = {021012},
  year    = {2022},
  doi     = {10.1115/1.4051604}
}

\clearpage
\appendix

\section{Finite-Element Implementation}
\label{app:fe_implementation}

\subsection{Component and impact model}

The datasets were generated using the component-level B-pillar side-impact model described in Section~\ref{subsec:datasets}. The simulations were performed in Virtual Performance Solution (VPS) using an explicit time-integration scheme. Each simulation covered \(30\)~ms.

The scaled B-pillar was discretised using triangular shell elements with a uniform thickness of \(1.6\)~mm and an average element size of approximately \(3\)~mm. The resulting meshes contained approximately \(4{,}000\) nodes and \(7{,}000\)--\(8{,}000\) shell elements, depending on the morphed geometry. Five through-thickness integration points were used for the shell elements.

The B-pillar was modelled as fully martensitic hot-stamped boron steel. The material model used a density of \(7.85\times10^{-6}\)~kg/mm\(^3\), a Young's modulus of \(205\)~GPa, and a Poisson's ratio of \(0.3\). Strain-rate dependence was represented by tabulated material curves defined at nominal strain rates of \(0.001\), \(5\), \(200\), and \(500\)~s\(^{-1}\). Details of this material can be found in \cite{li2026regunet}.

Impact was applied using a rigid cylindrical impactor with a diameter of \(40\)~mm and a prescribed mass of \(100\)~kg. Its translational motion in the impact direction was prescribed through the initial velocity. Surface-to-surface contact was defined between the impactor and the B-pillar with a friction coefficient of \(0.1\). Depending on the dataset, the initial impactor velocity ranged from \(3.5\) to \(4.0\)~m/s and its vertical impact position ranged from \(75\) to \(105\)~mm.

\subsection{Fixed and Elastic Support Conditions}
\label{app:fe_support_conditions}

The support conditions depend on the dataset considered. In Cases S and SL, the upper and lower connection regions of the B-pillar are fixed in all six degrees of freedom. In Case SLB, these fixed constraints are replaced by three independently varied elastic supports located at the top, bottom-left, and bottom-right connection regions. The elastic supports provide a simplified representation of the structural restraint transmitted by neighbouring components through the B-pillar connections.

The elastic supports are implemented in VPS using One-to-Many Constraint (OTMCO) interpolation formulation and one-dimensional spring elements. For each connection region $c\in\{\mathrm{T},\mathrm{BL},\mathrm{BR}\}$, an OTMCO constraint couples the corresponding mesh nodes to a reference node. OTMCO provides an interpolation constraint in which the motion of the reference node is determined from the collective motion of the coupled mesh nodes, and the resulting load is distributed back across the support region. This avoids applying the spring force to a single FE node and does not impose rigid-body motion on the coupled region.

A spring element is connected between each OTMCO reference node and a fully constrained anchor node. Motion of the support region is restricted in the remaining directions, while displacement along the spring axis is resisted by the elastic support. The spring response is linear and can be written as
\begin{equation}
    f^{\mathrm{sup}}_{c,t}=K^{\mathrm{sup}}_c\zeta_{c,t},
\end{equation}
where $\zeta_{c,t}$ is the relative displacement along the spring axis, $f^{\mathrm{sup}}_{c,t}$ is the corresponding support force, and $K^{\mathrm{sup}}_c$ is the prescribed support stiffness.

The spring elements use VPS material type 220 with a linear force--displacement curve. The stiffnesses of the top, bottom-left, and bottom-right supports are sampled independently between $1$ and $10$~kN/mm. Details can be found in \ref{app:dataset_design_space}. This produces both symmetric and asymmetric support configurations. The reaction forces at the three supports are recorded separately and used as scalar crashworthiness responses.

\subsection{Response extraction}

Nodal displacements, impactor motion, internal energy, contact force, and the three support reaction forces were extracted from each simulation. The learning datasets were sampled at intervals of \(1.5\)~ms. The first 12 response increments were retained, corresponding to the first \(18\)~ms of the impact event, during which the principal component deformation and peak force responses developed.

\section{Dataset Parameter Ranges and Design-Space Justification}
\label{app:dataset_design_space}

The reduced morphing ranges in Case SL are selected to compensate for the two additional loading variables and thereby maintain a comparable overall level of parameter variation under a fixed simulation budget. A direct comparison of hypervolumes would not be meaningful because Cases S and SL have different dimensionalities and combine parameters with different physical units. We therefore consider a simple nondimensional spread budget. Let $a_j$ denote the range width of parameter $j$, normalised by the largest range used for that parameter across the datasets, with $a_j=0$ for a fixed parameter. With equal weights, the spread budget is defined as
\begin{equation}
    B = \sum_{j=1}^{d} a_j^2 .
\end{equation}
This quantity is proportional to the expected squared distance from the centre of a uniformly sampled hyper-rectangular design space and therefore provides a unit-independent comparison of parameter spread. For Case S, the three shape parameters span their full ranges, giving $B_{\mathrm{S}}=1^2+1^2+1^2=3$. In Case SL, each shape range is halved and the two loading parameters span their full prescribed ranges, giving $B_{\mathrm{SL}}=3(0.5)^2+1^2+1^2=2.75$. Consequently, Case SL retains $2.75/3=91.7\%$ of the normalised squared spread of Case S, corresponding to $\sqrt{2.75/3}=95.7\%$ of its root-mean-square radius. The two cases therefore have similar overall normalised variability, while Case SL distributes this variability over a higher-dimensional combination of shape and loading parameters. This comparison concerns parameter spread rather than an exact equivalence of design-space volumes or physical sensitivities. Normalising variables before constructing a space-filling experimental design is also consistent with Latin-hypercube sampling practice for multidimensional computer experiments \cite{mckay1979latin}.

\begin{table}[htbp]
    \centering
    \caption{Parameter ranges used to generate the three B-pillar datasets. Cases S and SL use fixed constraints at the support regions.}
    \label{tab:dataset-parameter-ranges}
    \footnotesize
    \resizebox{\linewidth}{!}{%
    \begin{tabular}{@{}lcccccccc@{}}
        \toprule
        & \multicolumn{3}{c}{Shape range (\%)}
        & \multicolumn{2}{c}{Loading range}
        & \multicolumn{3}{c}{Support stiffness range (kN/mm)} \\
        \cmidrule(lr){2-4}\cmidrule(lr){5-6}\cmidrule(lr){7-9}
        Case
        & $X$ morph & $Y$ morph & $Z$ morph
        & Impact position (mm) & Impact velocity (m/s)
        & Top & Bottom left & Bottom right \\
        \midrule
        S
        & $[-5,5]$ & $[0,10]$ & $[-10,10]$
        & $90$ & $4.0$ & Fixed & Fixed & Fixed \\
        SL
        & $[-2.5,2.5]$ & $[0,5]$ & $[-5,5]$
        & $[75,105]$ & $[3.5,4.0]$ & Fixed & Fixed & Fixed \\
        SLB
        & $[-5,5]$ & $[0,10]$ & $[-10,10]$
        & $[75,105]$ & $[3.5,4.0]$ & $[1,10]$ & $[1,10]$ & $[1,10]$ \\
        \bottomrule
    \end{tabular}%
    }
\end{table}

\section{Training and Validation Error Distributions}
\label{app:train_validation_boxplots}

\begin{figure}[H]
    \centering
    \includegraphics[width=0.95\textwidth]{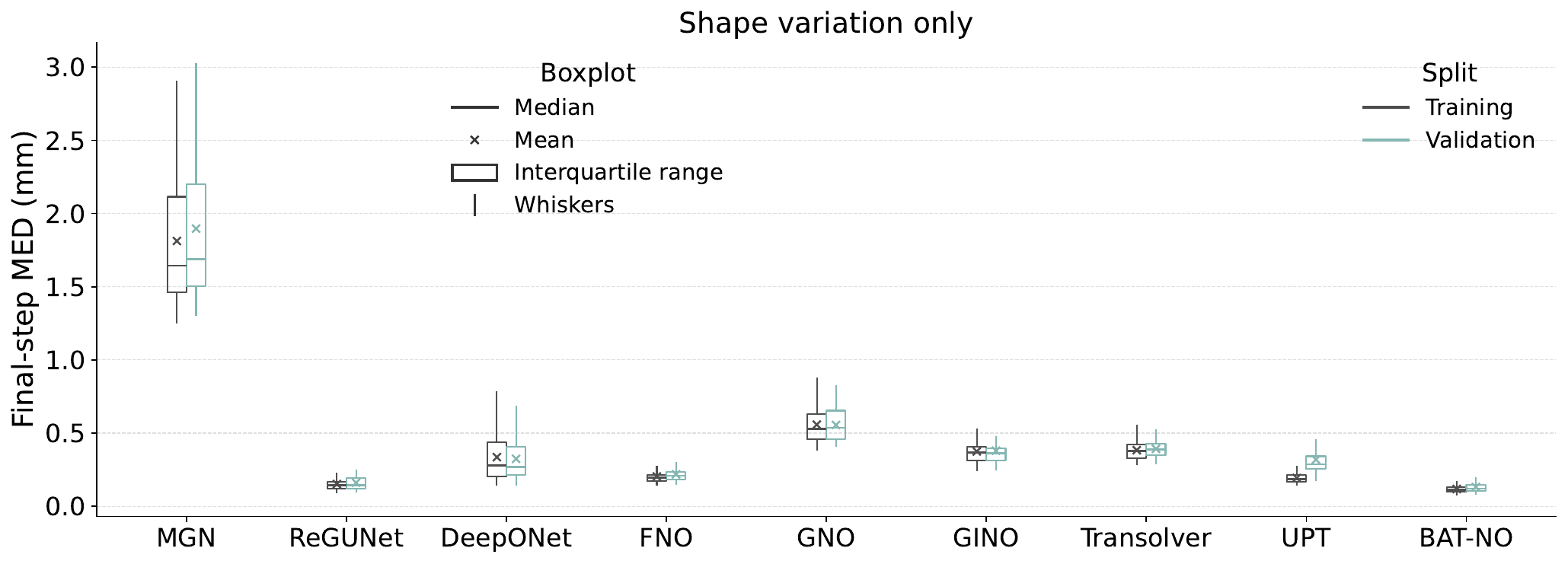}
    \caption{Box plots comparing the training and validation final-step MED values of the baseline models for Case S, in which only the B-pillar shape is varied.}
    \label{fig:boxplot_case_s}
\end{figure}

Figure~\ref{fig:boxplot_case_s} shows the error distributions for the shape-only case. Most operator-based models generalise consistently from the training set to the validation set, indicating that geometric variation alone can be learned with limited distribution shift. ReGUNet and FNO also show compact validation distributions, while MGN and GNO have larger errors. ReGUNet performs particularly well in this shape-variation setting, suggesting that its hierarchical graph U-Net architecture is particularly sensitive to geometry changes and can capture the resulting multiscale deformation features. BAT-NO gives the lowest validation error, with a mean final-step MED of $0.132$~mm. This suggests that, for this case, latent-grid processing and recurrent or operator-based representations are more effective than relying only on local graph propagation.

\begin{figure}[H]
    \centering
    \includegraphics[width=0.95\textwidth]{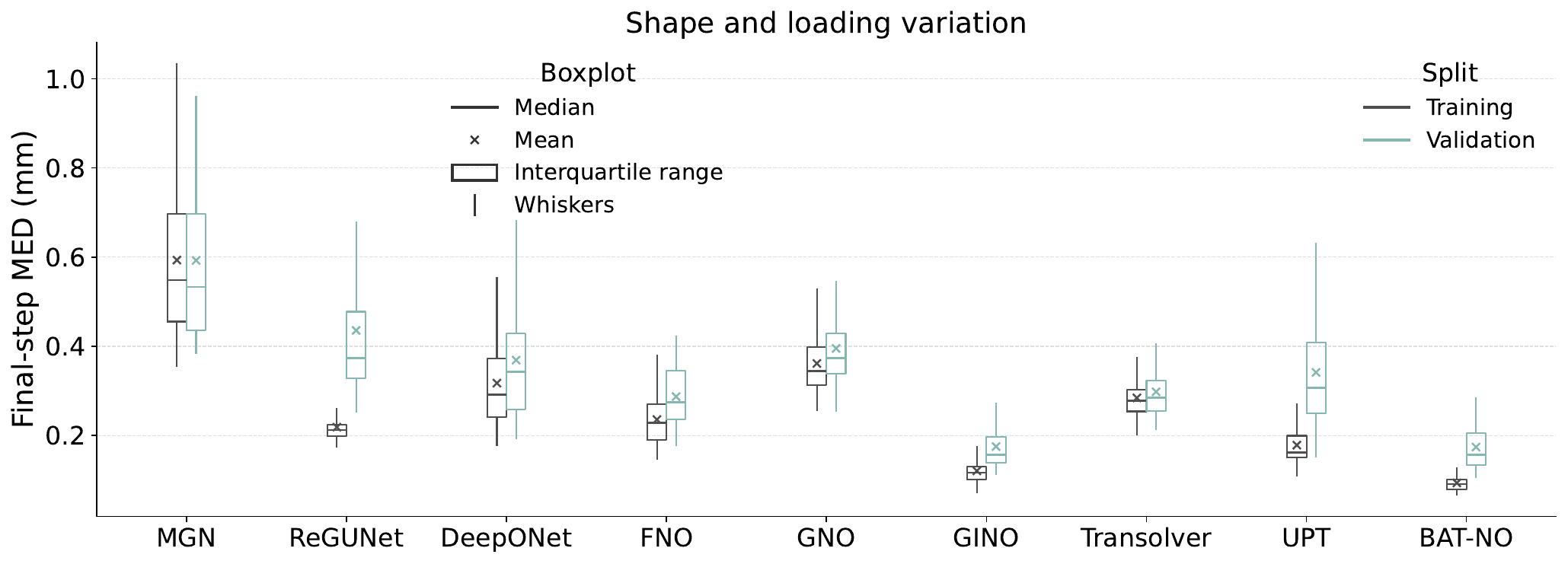}
    \caption{Box plots comparing the training and validation final-step MED values of the baseline models for Case SL, in which the B-pillar shape and loading conditions are varied.}
    \label{fig:boxplot_case_sl}
\end{figure}

Figure~\ref{fig:boxplot_case_sl} reports the results when both shape and loading conditions are varied. The shape-variation ranges are reduced relative to Case S, while variations in impactor position and initial velocity are introduced. Several models achieve lower errors in Case SL than in Case S, indicating that the reduced geometric variation may offset the additional difficulty introduced by loading variation. Because the two cases use different geometry ranges, their errors do not isolate the effect of loading variation alone. ReGUNet loses its strong relative position from the shape-only case, whereas GINO and BAT-NO perform comparably, with mean validation errors of $0.175$ and $0.174$~mm, respectively.

\begin{figure}[H]
    \centering
    \includegraphics[width=0.95\textwidth]{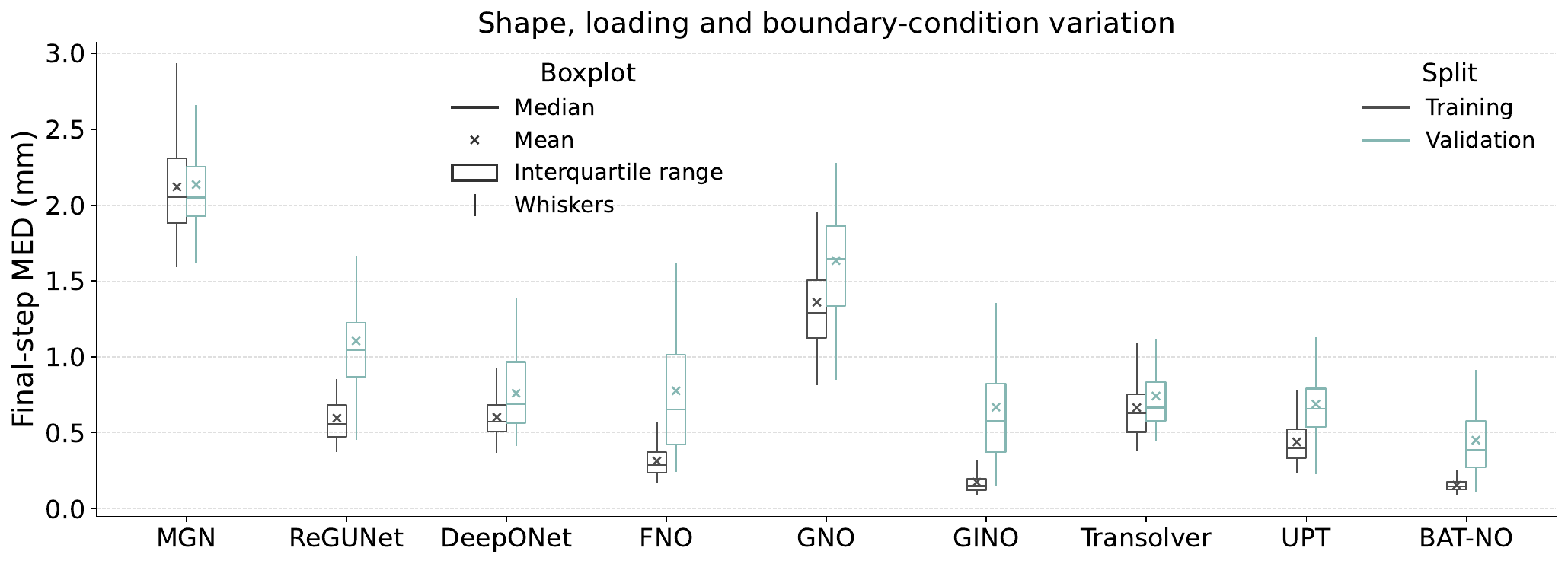}
    \caption{Box plots comparing the training and validation final-step MED values of the baseline models for Case SLB, in which the B-pillar shape, loading conditions, and three spring-based boundary conditions are varied.}
    \label{fig:boxplot_case_slb}
\end{figure}

Figure~\ref{fig:boxplot_case_slb} shows the most challenging setting, where shape, loading, and spring-based boundary conditions are varied simultaneously. All models exhibit larger validation errors and wider distributions, confirming that boundary and loading-condition variation introduces a stronger source of response variability. BAT-NO obtains the lowest mean validation error of $0.451$~mm, followed by GINO. The improvement of BAT-NO is consistent with its explicit boundary-condition encoding, which combines local condition projection with slice-based global communication. The larger training-validation gaps in this case also show that generalisation across boundary-condition changes remains the dominant challenge among the three case studies.

\section{Hyperparameter Tuning}
\label{app:hyperparameter_tuning}

The hyperparameter study was conducted as a staged one-factor screening process. Each stage varied one parameter while keeping the current reference setting fixed. The reported error is the mean final-step MED on the validation set. Each screening configuration was trained for 250 epochs to reduce the computational cost of the search. After selecting the final hyperparameters, the tuned model was trained for 1000 epochs for full validation and evaluation.

For the learning rate, Table~\ref{tab:hp-learning-rate} shows that very small learning rates underfit within the fixed training budget, while larger values lead to less stable optimisation. The best screening result is obtained at $3\times10^{-4}$, which provides the lowest validation error among the tested candidates. This value was therefore selected for the subsequent experiments. Table~\ref{tab:hp-lr-schedule} compares a constant learning rate with a cosine schedule using warm-up. The constant schedule gives lower validation error in this setting. This suggests that, for the current training length and dataset size, maintaining the learning rate is more effective than gradually reducing it. The constant schedule was therefore used in the final training. 

\begin{table}[htbp]
\centering
\begin{minipage}[t]{0.48\textwidth}
\vspace{0pt}
\centering
\caption{Learning-rate screening.}
\label{tab:hp-learning-rate}
\footnotesize
\begin{tabularx}{\linewidth}{@{}Xr@{}}
\toprule
Learning rate & Final-step MED (mm) \\
\midrule
$1\times10^{-4}$ & 0.788 \\
$2\times10^{-4}$ & 0.728 \\
$\mathbf{3\times10^{-4}}$ & \textbf{0.529} \\
$4\times10^{-4}$ & 0.586 \\
$6\times10^{-4}$ & 0.571 \\
$8\times10^{-4}$ & 0.621 \\
\bottomrule
\end{tabularx}
\end{minipage}
\hfill
\begin{minipage}[t]{0.48\textwidth}
\vspace{0pt}
\centering
\caption{Learning-rate-schedule screening.}
\label{tab:hp-lr-schedule}
\footnotesize
\begin{tabularx}{\linewidth}{@{}Xr@{}}
\toprule
Schedule & Final-step MED (mm) \\
\midrule
\textbf{Constant} & \textbf{0.529} \\
Cosine with 10-epoch warm-up & 0.686 \\
\bottomrule
\end{tabularx}
\end{minipage}
\end{table}

The scalar-loss weight controls the relative contribution of scalar-response prediction to the total loss. As shown in Table~\ref{tab:hp-scalar-loss-weight}, a small weight of 0.05 gives weaker field accuracy, while increasing the weight to 0.20 also degrades the result. This indicates that scalar supervision is useful, but should not dominate the displacement-field objective. A scalar-loss weight of 0.10 gives the best validation error and was selected. Table~\ref{tab:hp-batch-size} compares batch sizes of 2 and 4. A batch size of 8 was also attempted but exceeded the available GPU-memory limit and is therefore not included. Increasing the batch size worsens the validation error, possibly because the smaller batch introduces useful stochasticity during optimisation and because the effective number of parameter updates per epoch is larger. Batch size 2 also remains compatible with the GPU-memory limit for the larger grid-resolution tests. Therefore, batch size 2 was retained.

\begin{table}[htbp]
\centering
\begin{minipage}[t]{0.48\textwidth}
\vspace{0pt}
\centering
\caption{Scalar-loss-weight screening.}
\label{tab:hp-scalar-loss-weight}
\footnotesize
\begin{tabularx}{\linewidth}{@{}Xr@{}}
\toprule
Scalar-loss weight & Final-step MED (mm) \\
\midrule
0.05 & 0.731 \\
\textbf{0.10} & \textbf{0.529} \\
0.20 & 0.567 \\
\bottomrule
\end{tabularx}
\end{minipage}
\hfill
\begin{minipage}[t]{0.48\textwidth}
\vspace{0pt}
\centering
\caption{Batch-size screening.}
\label{tab:hp-batch-size}
\footnotesize
\begin{tabularx}{\linewidth}{@{}Xr@{}}
\toprule
Batch size & Final-step MED (mm) \\
\midrule
\textbf{2} & \textbf{0.529} \\
4 & 0.645 \\
\bottomrule
\end{tabularx}
\end{minipage}
\end{table}

Table~\ref{tab:hp-grid-resolution} summarises the latent-grid-resolution screening. As discussed in Section~\ref{subsec:final_model_configuration}, the $20^3$ grid was selected as the cost-aware balance between validation accuracy and computational requirements. The number of Fourier modes controls the spectral bandwidth retained in each FNO layer. Table~\ref{tab:hp-fourier-modes} shows that four modes give the lowest validation error. Using fewer modes limits the representation of spatial variation, while using more modes does not improve the result and may make optimisation less stable for the available data. Four Fourier modes were therefore retained. Note that the initial grid screening selected a $16^3$ grid, giving an error of $0.480$~mm. The Fourier-mode screening was conducted using this $16^3$ configuration, which explains its reference value of $0.480$~mm. The extended grid screening subsequently selected the $20^3$ cost-aware configuration, with an error of $0.445$~mm. This configuration was inherited by the FNO-depth, slice-count, and attention-head screenings.

\begin{table}[htbp]
\centering
\begin{minipage}[t]{0.48\textwidth}
\vspace{0pt}
\centering
\caption{Latent-grid-resolution screening. The $20^3$ grid is the selected cost-aware operating point.}
\label{tab:hp-grid-resolution}
\footnotesize
\begin{tabularx}{\linewidth}{@{}Xr@{}}
\toprule
Grid resolution & Final-step MED (mm) \\
\midrule
$8^3$ & 0.529 \\
$12^3$ & 0.554 \\
$16^3$ & 0.480 \\
$\mathbf{20^3}$ & \textbf{0.445} \\
$24^3$ & 0.413 \\
$28^3$ & GPU memory limit exceeded \\
\bottomrule
\end{tabularx}
\end{minipage}
\hfill
\begin{minipage}[t]{0.48\textwidth}
\vspace{0pt}
\centering
\caption{Fourier-mode screening.}
\label{tab:hp-fourier-modes}
\footnotesize
\begin{tabularx}{\linewidth}{@{}Xr@{}}
\toprule
Fourier modes & Final-step MED (mm) \\
\midrule
3 & 0.611 \\
\textbf{4} & \textbf{0.480} \\
6 & 0.693 \\
\bottomrule
\end{tabularx}
\end{minipage}
\end{table}

Tables~\ref{tab:hp-fno-layers} and~\ref{tab:hp-slice-count} summarise the FNO-depth and slice-count screenings, respectively. As discussed in Section~\ref{subsec:final_model_configuration}, the final model uses 3 FNO layers and 16 boundary-condition slices.

\begin{table}[htbp]
\centering
\begin{minipage}[t]{0.48\textwidth}
\vspace{0pt}
\centering
\caption{FNO-depth screening.}
\label{tab:hp-fno-layers}
\footnotesize
\begin{tabularx}{\linewidth}{@{}Xr@{}}
\toprule
FNO layers & Final-step MED (mm) \\
\midrule
2 & 0.519 \\
\textbf{3} & \textbf{0.445} \\
4 & 0.584 \\
\bottomrule
\end{tabularx}
\end{minipage}
\hfill
\begin{minipage}[t]{0.48\textwidth}
\vspace{0pt}
\centering
\caption{Slice-count screening. Sixteen slices are retained as the cost-aware selection.}
\label{tab:hp-slice-count}
\footnotesize
\begin{tabularx}{\linewidth}{@{}Xr@{}}
\toprule
Number of slices & Final-step MED (mm) \\
\midrule
8 & 0.520 \\
\textbf{16} & \textbf{0.445} \\
32 & 0.442 \\
64 & 0.477 \\
\bottomrule
\end{tabularx}
\end{minipage}
\end{table}

The number of attention heads controls how many independent slice-assignment and attention subspaces are used. Table~\ref{tab:hp-attention-heads} shows that four heads give the best result among the main candidates. Two heads are too restrictive, while eight and sixteen heads do not improve the validation error. This indicates that a moderate number of heads is enough to capture distinct boundary-condition interactions without adding unnecessary complexity. Four heads were selected. The pre-grid message-passing depth controls how much local mesh information is propagated before the mesh-to-grid projection. Table~\ref{tab:hp-pre-grid-mp} shows that increasing the depth from one to four message-passing steps reduces the validation error. This supports the role of local graph processing before global grid-based operator learning. The eight-step model exceeds the available GPU memory. Four pre-grid message-passing steps were therefore selected.

\begin{table}[htbp]
\centering
\begin{minipage}[t]{0.48\textwidth}
\vspace{0pt}
\centering
\caption{Attention-head screening.}
\label{tab:hp-attention-heads}
\footnotesize
\begin{tabularx}{\linewidth}{@{}Xr@{}}
\toprule
Attention heads & Final-step MED (mm) \\
\midrule
2 & 0.759 \\
\textbf{4} & \textbf{0.445} \\
8 & 0.513 \\
16 & 0.465 \\
\bottomrule
\end{tabularx}
\end{minipage}
\hfill
\begin{minipage}[t]{0.48\textwidth}
\vspace{0pt}
\centering
\caption{Pre-grid message-passing-depth screening.}
\label{tab:hp-pre-grid-mp}
\footnotesize
\begin{tabularx}{\linewidth}{@{}Xr@{}}
\toprule
MP steps & Final-step MED (mm) \\
\midrule
1 & 0.445 \\
2 & 0.456 \\
\textbf{4} & \textbf{0.412} \\
8 & GPU memory limit exceeded \\
\bottomrule
\end{tabularx}
\end{minipage}
\end{table}

The slice-transformer depth controls the number of self-attention layers applied to the boundary-condition slice tokens. Table~\ref{tab:hp-slice-transformer-depth} shows that one transformer layer gives the best validation error. Adding more layers increases the error, suggesting that deeper token interaction is not required for the current dataset and may over-parameterise the boundary-condition pathway. One slice-transformer layer was retained. Attention dropout was tested as a regularisation mechanism for the slice-attention pathway. As shown in Table~\ref{tab:hp-attention-dropout}, adding dropout increases the validation error. This suggests that the current model does not benefit from stochastic removal of attention weights, possibly because the boundary-condition token pathway is already compact. The final model therefore uses no attention dropout.

\begin{table}[htbp]
\centering
\begin{minipage}[t]{0.48\textwidth}
\vspace{0pt}
\centering
\caption{Slice-transformer-depth screening.}
\label{tab:hp-slice-transformer-depth}
\footnotesize
\begin{tabularx}{\linewidth}{@{}Xr@{}}
\toprule
Transformer layers & Final-step MED (mm) \\
\midrule
\textbf{1} & \textbf{0.412} \\
2 & 0.472 \\
3 & 0.486 \\
\bottomrule
\end{tabularx}
\end{minipage}
\hfill
\begin{minipage}[t]{0.48\textwidth}
\vspace{0pt}
\centering
\caption{Attention-dropout screening.}
\label{tab:hp-attention-dropout}
\footnotesize
\begin{tabularx}{\linewidth}{@{}Xr@{}}
\toprule
Attention dropout & Final-step MED (mm) \\
\midrule
\textbf{0} & \textbf{0.412} \\
0.05 & 0.422 \\
0.10 & 0.443 \\
\bottomrule
\end{tabularx}
\end{minipage}
\end{table}

Weight decay was also tested as a regularisation mechanism. Table~\ref{tab:hp-weight-decay} shows that $1\times10^{-6}$ gives almost the same result as zero weight decay, while $1\times10^{-5}$ substantially worsens the validation error. This indicates that stronger parameter shrinkage is harmful for the current model. Zero weight decay was selected. The maximum predicted-feedback probability controls the degree of scheduled autoregressive training. A value of zero corresponds to teacher-forced training, while larger values increase the probability of feeding model predictions back into the next step during training. Table~\ref{tab:hp-predicted-feedback} shows that predicted feedback improves the validation error compared with pure teacher forcing, with the best screening result at 0.75. 

\begin{table}[htbp]
\centering
\begin{minipage}[t]{0.48\textwidth}
\vspace{0pt}
\centering
\caption{Weight-decay screening.}
\label{tab:hp-weight-decay}
\footnotesize
\begin{tabularx}{\linewidth}{@{}Xr@{}}
\toprule
Weight decay & Final-step MED (mm) \\
\midrule
\textbf{0} & \textbf{0.412} \\
$1\times10^{-6}$ & 0.413 \\
$1\times10^{-5}$ & 0.568 \\
\bottomrule
\end{tabularx}
\end{minipage}
\hfill
\begin{minipage}[t]{0.48\textwidth}
\vspace{0pt}
\centering
\caption{Maximum predicted-feedback-probability screening.}
\label{tab:hp-predicted-feedback}
\footnotesize
\begin{tabularx}{\linewidth}{@{}Xr@{}}
\toprule
Maximum feedback probability & Final-step MED (mm) \\
\midrule
0.00 & 0.412 \\
0.25 & 0.434 \\
0.50 & 0.335 \\
\textbf{0.75} & \textbf{0.331} \\
1.00 & 0.343 \\
\bottomrule
\end{tabularx}
\end{minipage}
\end{table}

\begin{figure}[p]
\centering
\begin{minipage}[t]{0.48\textwidth}
    \centering
    \includegraphics[width=\linewidth]{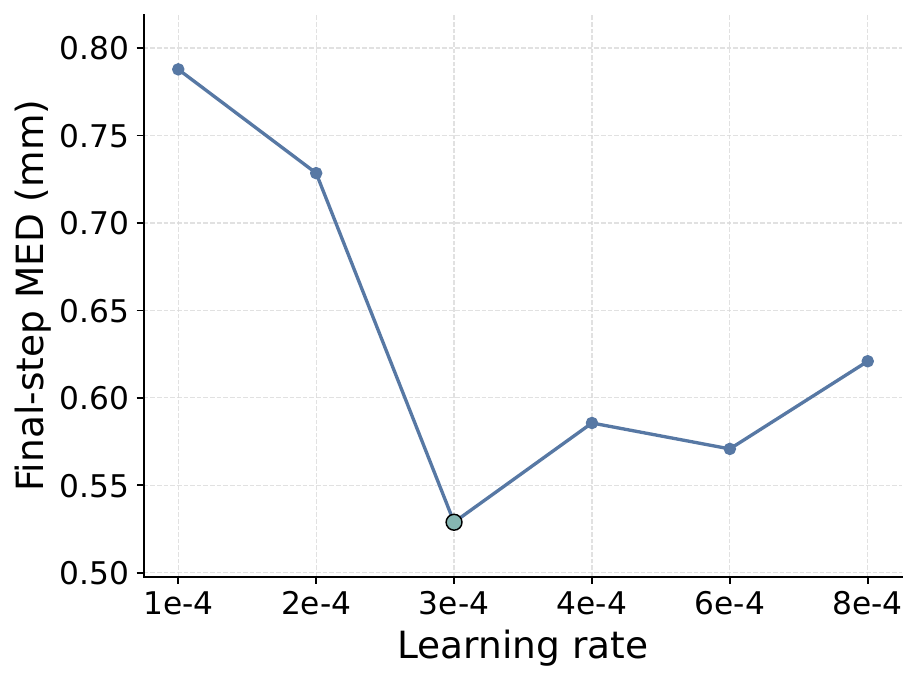}
    \small (a) Learning rate
\end{minipage}
\hfill
\begin{minipage}[t]{0.48\textwidth}
    \centering
    \includegraphics[width=\linewidth]{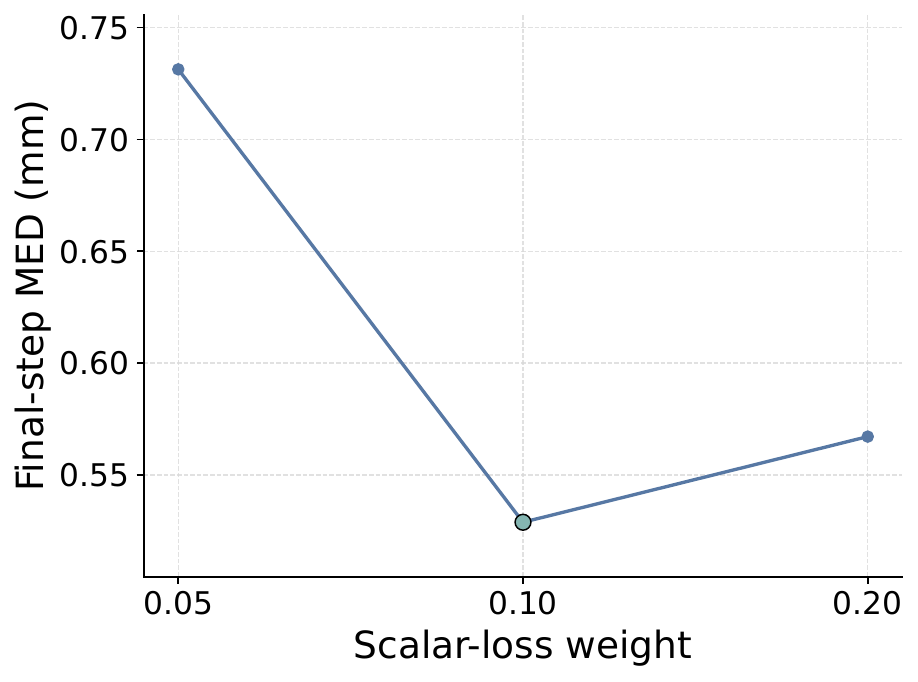}
    \small (b) Scalar-loss weight
\end{minipage}

\vspace{0.8em}

\begin{minipage}[t]{0.48\textwidth}
    \centering
    \includegraphics[width=\linewidth]{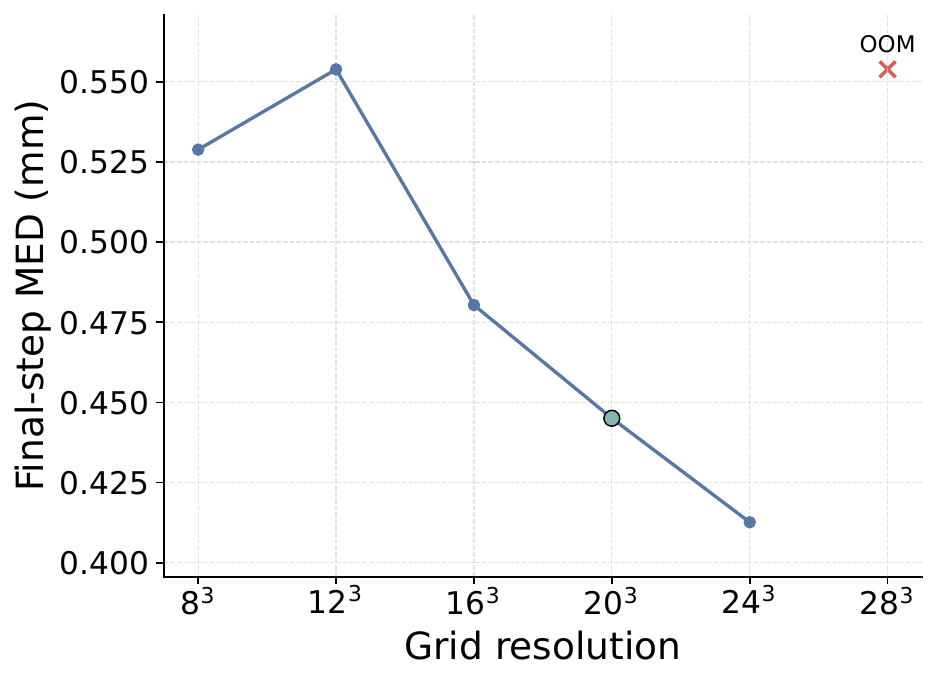}
    \small (c) Latent-grid resolution
\end{minipage}
\hfill
\begin{minipage}[t]{0.48\textwidth}
    \centering
    \includegraphics[width=\linewidth]{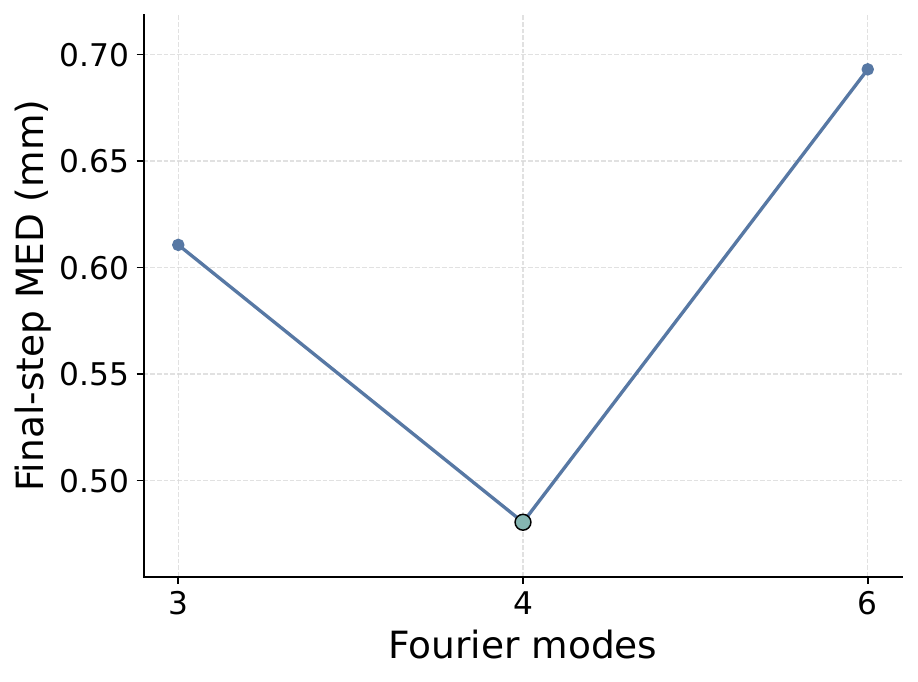}
    \small (d) Fourier modes
\end{minipage}

\vspace{0.8em}

\begin{minipage}[t]{0.48\textwidth}
    \centering
    \includegraphics[width=\linewidth]{figures/hp_figures/hyperparameter_FNO_layer.pdf}
    \small (e) FNO depth
\end{minipage}
\hfill
\begin{minipage}[t]{0.48\textwidth}
    \centering
    \includegraphics[width=\linewidth]{figures/hp_figures/hyperparameter_slice_count.pdf}
    \small (f) Slice count
\end{minipage}
\caption{Validation final-step MED for the first six BAT-NO hyperparameter screenings: (a) learning rate, (b) scalar-loss weight, (c) latent-grid resolution, (d) Fourier modes, (e) FNO depth, and (f) slice count.}
\label{fig:hp-screening-panels-1}
\end{figure}

\begin{figure}[p]
\centering
\begin{minipage}[t]{0.48\textwidth}
    \centering
    \includegraphics[width=\linewidth]{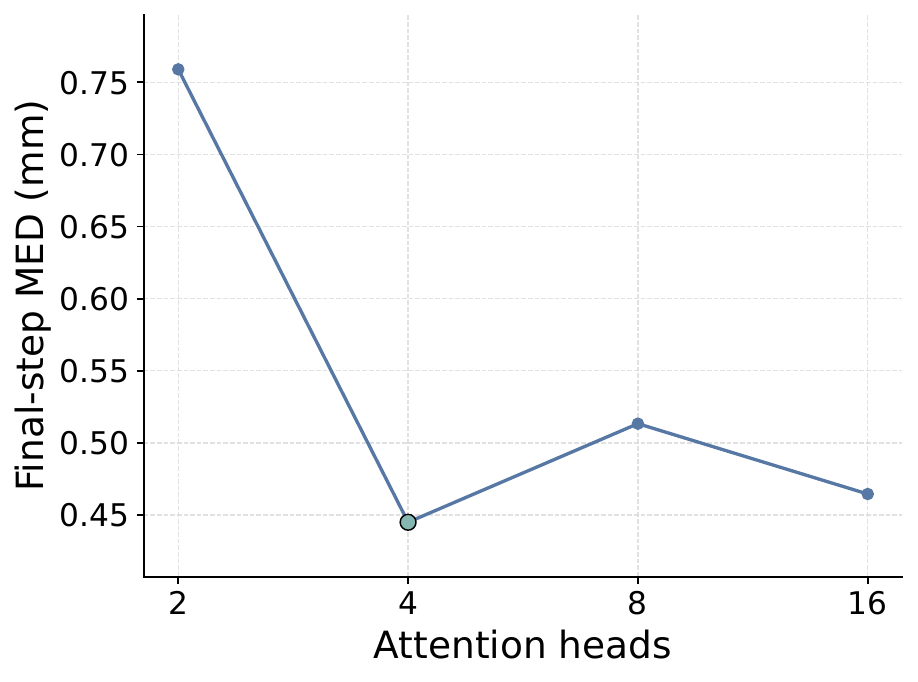}
    \small (a) Attention heads
\end{minipage}
\hfill
\begin{minipage}[t]{0.48\textwidth}
    \centering
    \includegraphics[width=\linewidth]{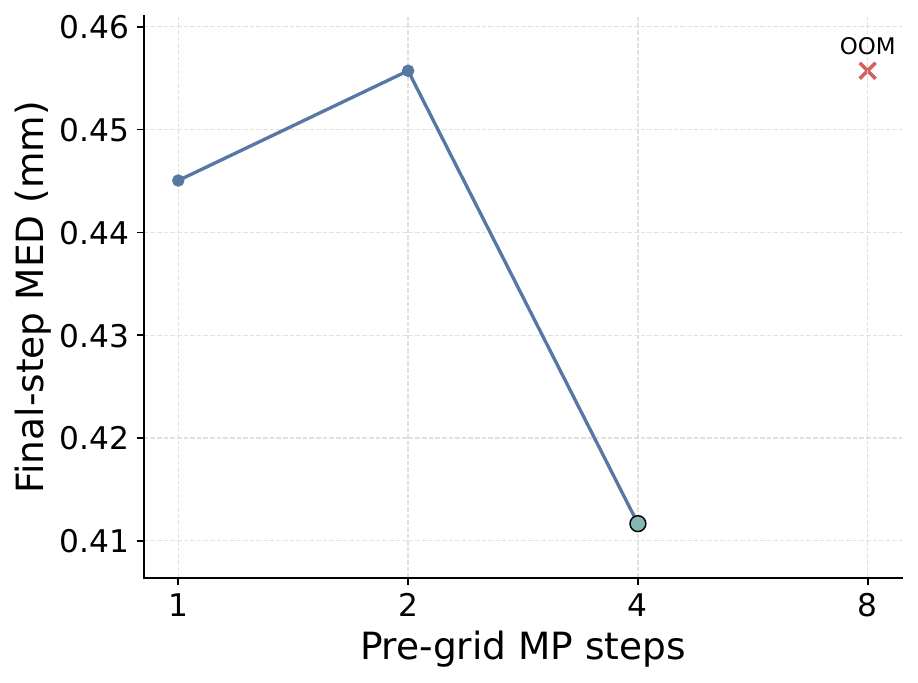}
    \small (b) Pre-grid message-passing steps
\end{minipage}

\vspace{0.8em}

\begin{minipage}[t]{0.48\textwidth}
    \centering
    \includegraphics[width=\linewidth]{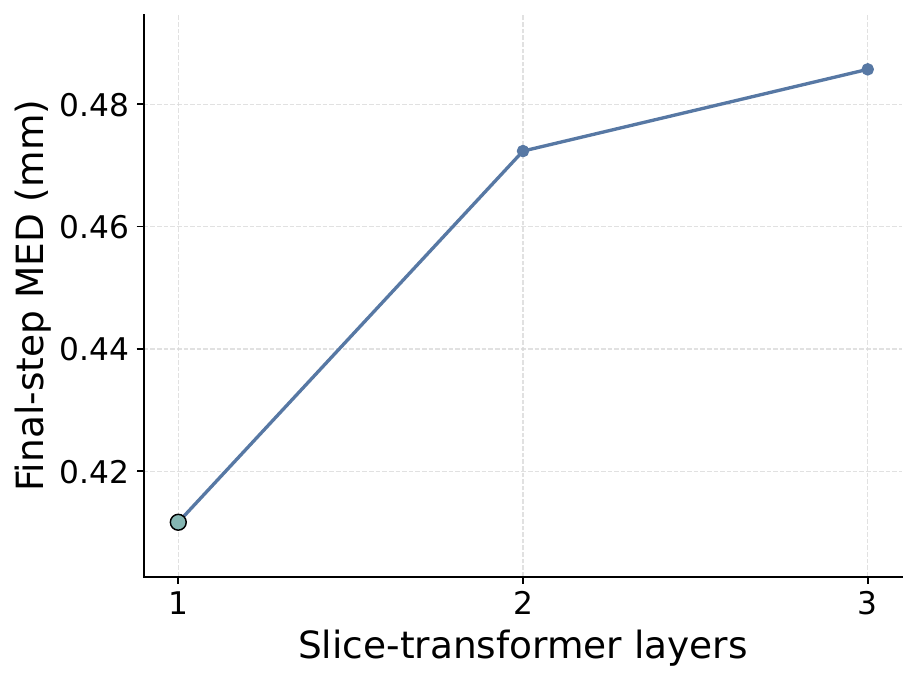}
    \small (c) Slice-transformer depth
\end{minipage}
\hfill
\begin{minipage}[t]{0.48\textwidth}
    \centering
    \includegraphics[width=\linewidth]{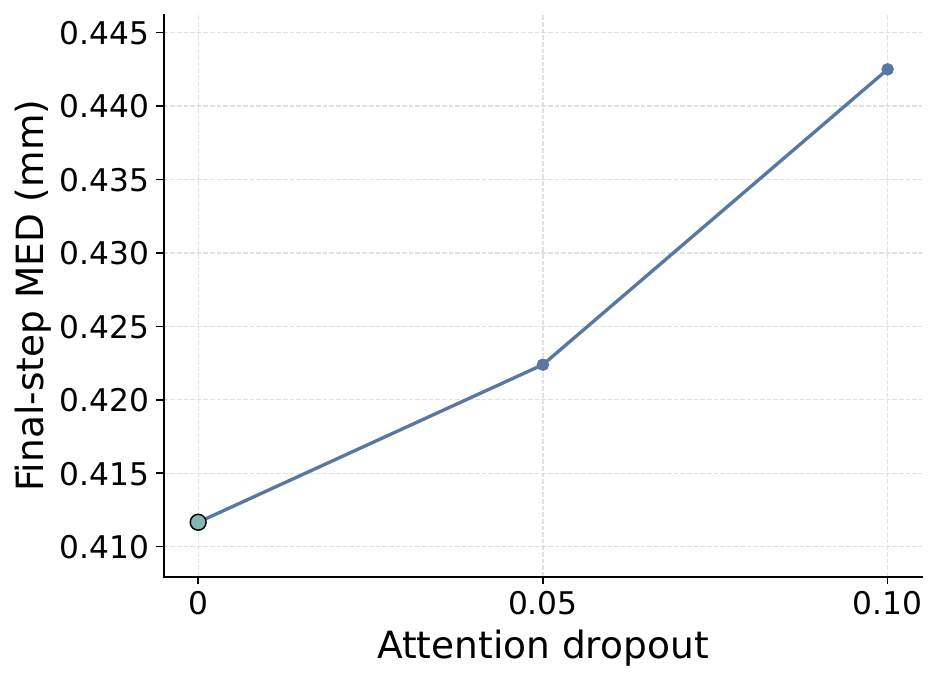}
    \small (d) Attention dropout
\end{minipage}

\vspace{0.8em}

\begin{minipage}[t]{0.48\textwidth}
    \centering
    \includegraphics[width=\linewidth]{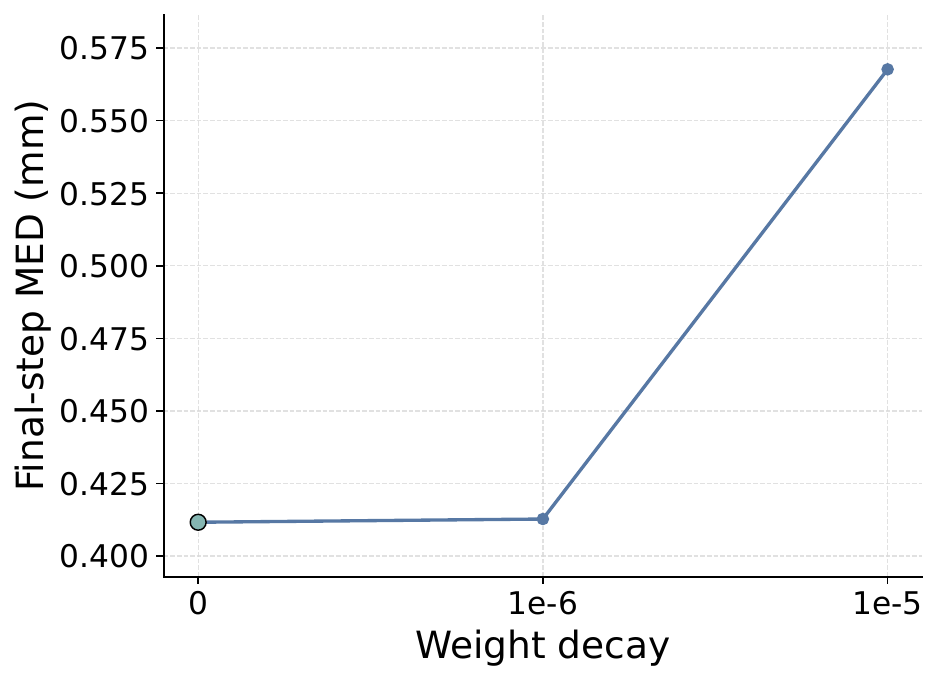}
    \small (e) Weight decay
\end{minipage}
\hfill
\begin{minipage}[t]{0.48\textwidth}
    \centering
    \includegraphics[width=\linewidth]{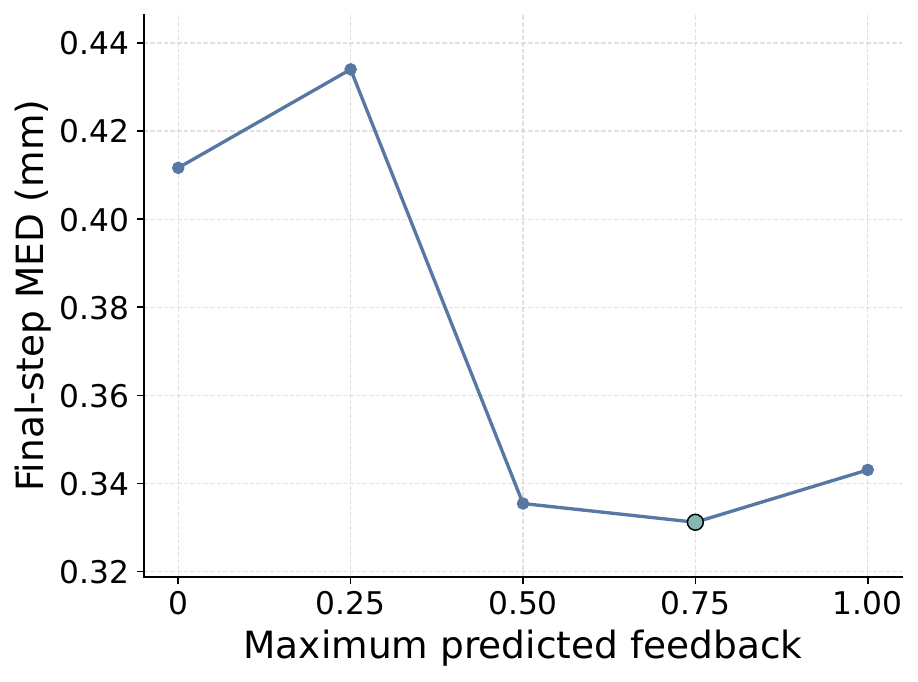}
    \small (f) Maximum predicted-feedback probability
\end{minipage}
\caption{Validation final-step MED for the remaining BAT-NO hyperparameter screenings: (a) attention heads, (b) pre-grid message-passing steps, (c) slice-transformer depth, (d) attention dropout, (e) weight decay, and (f) maximum predicted-feedback probability.}
\label{fig:hp-screening-panels-2}
\end{figure}

Figures~\ref{fig:hp-screening-panels-1} and~\ref{fig:hp-screening-panels-2} visualise the relationship between each screened hyperparameter and the validation error, providing a more direct comparison of the candidate settings than the tabulated values alone.

The final BAT-NO model was trained using the settings summarised in Table~\ref{tab:training_details}. Adam optimisation was used with a constant learning rate of $3\times10^{-4}$, zero weight decay, and a batch size of 2. The model was trained for 1000 epochs. To reduce the mismatch between teacher-forced training and autoregressive evaluation, scheduled sampling was applied during training. The probability of feeding the model prediction back into the next time step was increased after a 20-epoch warm-up and ramped over 100 epochs to a maximum value of 0.75. Gradient clipping with a maximum norm of 1.0 was used to improve optimisation stability. The total loss combines the MSE of the normalised incremental-displacement field and the MSE of the normalised scalar responses, with the scalar term weighted by $\lambda_{\mathrm{scalar}}=0.1$.

\begin{table}[htbp]
\centering
\caption{Training and evaluation settings used for the final BAT-NO model.}
\label{tab:training_details}
\begin{tabularx}{\linewidth}{@{}lX@{}}
\hline
Setting & Value \\
\hline
Optimiser & Adam \\
Learning rate & $3\times10^{-4}$ \\
Learning-rate schedule & Constant \\
Weight decay & $0$ \\
Batch size & $2$ \\
Training epochs & $1000$ \\
Temporal input strategy & Scheduled sampling \\
Maximum predicted-feedback probability & $0.75$ \\
Feedback warm-up & $20$ epochs \\
Feedback ramp & $100$ epochs \\
Gradient clipping & Max norm $=1.0$ \\
Scalar-loss coefficient $\lambda_{\mathrm{scalar}}$ & $0.1$ \\
\hline
\end{tabularx}
\end{table}

\section{Full-Trajectory Field Visualisation on the Test Set}
\label{app:full_trajectory_visualisation}

\begin{figure}[htbp]
    \centering
    \includegraphics[width=\textwidth]{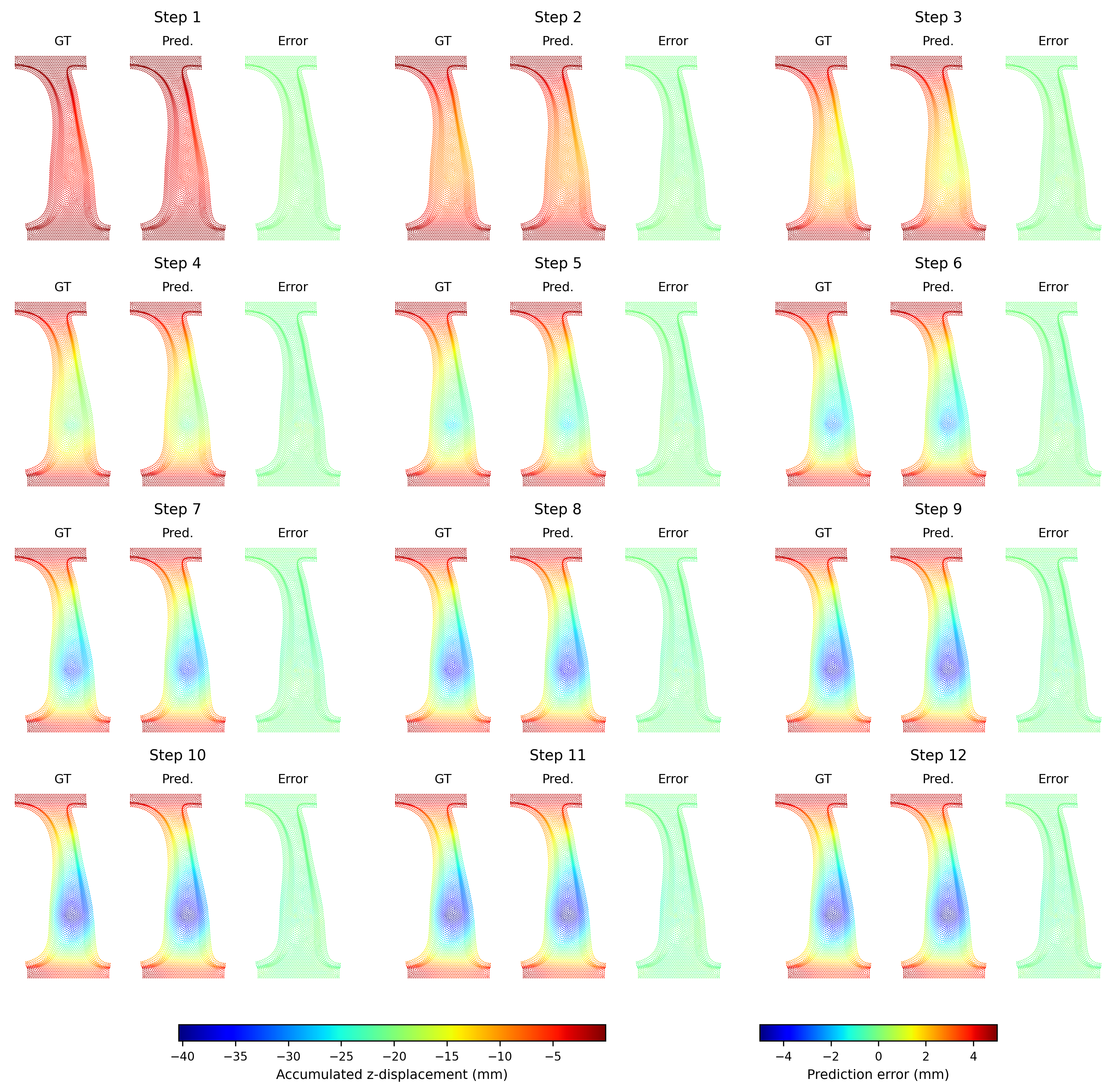}
    \caption{
    Full-trajectory visualisation of the accumulated $z$-displacement field for the median-error test sample. 
    Each time step shows the ground-truth field, BAT-NO prediction, and prediction error. 
    The error colour range is fixed to $\pm 5$~mm for all time steps.
    }
    \label{fig:test_median_rollout_z_field}
\end{figure}

Figure~\ref{fig:test_median_rollout_z_field} shows the full autoregressive rollout of the accumulated $z$-displacement field for the median-error test sample. The sample is selected from the 300-sample test set as the case closest to the median final-step MED of the tuned BAT-NO model. The figure reports the ground-truth field, predicted field, and prediction error at all 12 rollout steps.

The prediction remains close to the ground-truth displacement pattern throughout the rollout. The main deformation mode is captured from the early increments and remains consistent as the accumulated displacement increases. The error maps show that most nodal errors remain close to zero, with no strong spatial drift over time. The final-step MED of this test sample is $0.233$~mm, which is close to the median test-set behaviour. Therefore, the visualisation provides a representative qualitative example of the model performance on unseen simulations rather than a best-case result.

\end{document}